\documentclass[runningheads]{llncs}

\usepackage[mobile]{eccv}

\usepackage{pgfplots}
\pgfplotsset{compat=1.18}

\usepackage{eccvabbrv}

\usepackage{graphicx}
\usepackage{booktabs}
\usepackage{makecell}
\usepackage{pifont}
\usepackage{paralist}
\usepackage[percent]{overpic}
\usepackage{xcolor}
\usepackage{subcaption}
\usepackage{wrapfig}
\usepackage[table]{xcolor}
\usepackage{tcolorbox}
\usepackage{float}
\usepackage{placeins}
\tcbuselibrary{breakable}

\newtcolorbox{mybox}[1]{
    breakable,
    title=#1,
    colback=blue!5,
    colframe=blue!60,
    coltitle=white,
    colbacktitle=blue!60,
    fonttitle=\bfseries,
    boxrule=0.8pt,
    arc=3pt,
    left=4pt,
    right=4pt,
    top=4pt,
    bottom=4pt
}

\usepackage[accsupp]{axessibility}  

\usepackage{hyperref}

\usepackage{orcidlink}

\begin{document}

\title{Watching Movies Like a Human:
Egocentric Emotion Understanding for Embodied Companions} 
    
\titlerunning{Egocentric Emotion Understanding}

\author{
Dong Ze\inst{1}$^{*}$ \and
Hao Shi\inst{2,3}$^{*}$ \and
Zejia Gao\inst{4} \and
Zhonghua Yi\inst{2} \and
Kaiwei Wang\inst{2}$^{\dagger}$ \and
Lin Wang\inst{1}$^{\dagger}$
}

\authorrunning{Z.Dong et al.}

\institute{
$^{1}$ Nanyang Technological University, Singapore \hspace{1em}
\inst{2} Zhejiang University, China \\
\inst{3} Ant Group, China \hspace{1em}
\inst{4} MirrorMe, China
}

\maketitle
\begingroup
\renewcommand\thefootnote{}
\footnotetext{* Equal contribution.}
\footnotetext{$\dagger$ Corresponding authors: linwang@ntu.edu.sg, kaiweiwang@zju.edu.cn}
\endgroup

\begin{abstract}
Embodied robotic agents often perceive movies through an egocentric screen-view interface rather than native cinematic footage, introducing domain shifts such as viewpoint distortion, scale variation, illumination changes, and environmental interference. However, existing research on movie emotion understanding is almost exclusively conducted on cinematic footage, limiting cross-domain generalization to real-world viewing scenarios. 
To bridge this gap, we introduce \textit{EgoScreen-Emotion} (ESE), the {first} benchmark dataset for egocentric screen-view movie emotion understanding. ESE contains 224 movie trailers captured under controlled egocentric screen-view conditions, producing 28,667 temporally aligned key-frames annotated by multiple raters with a confidence-aware multi-label protocol to address emotional ambiguity. 
We further build a multimodal long-context emotion reasoning framework that models temporal visual evidence, narrative summaries, compressed historical context, and audio cues. Cross-domain experiments reveal a severe domain gap: models trained on cinematic footage drop from 27.99 to 16.69 Macro-F1 when evaluated on realistic egocentric screen-view observations. Training on ESE substantially improves robustness under realistic viewing conditions. Our approach achieves competitive performance compared with strong closed-source multimodal models, highlighting the importance of domain-specific data and long-context multimodal reasoning. 

\end{abstract}

\section{Introduction}
\label{sec:intro}
\begin{figure}[t]
    \centering
    \includegraphics[width=\linewidth, trim=1cm 2.25cm 1cm 2.2cm, clip]{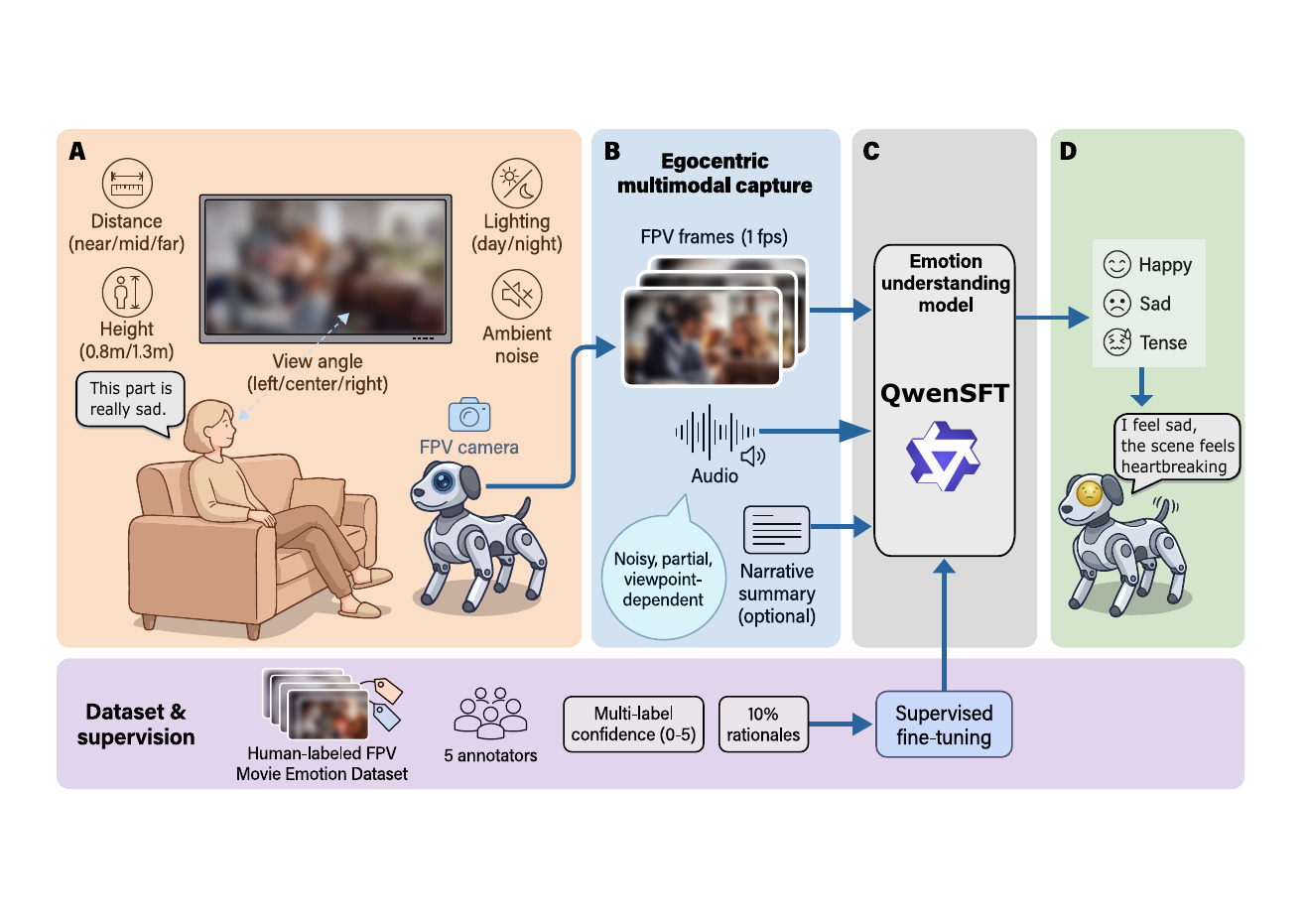}
    \caption{
    \textbf{Overview of our embodied affective reasoning framework.} An agent captures egocentric movie frames and multimodal context in real-world viewing scenarios. A fine-tuned Qwen-based model processes these inputs to predict intrinsic viewer emotions and generate empathetic feedback, supervised by structured human annotations from our dataset.
    }
    \label{fig:overview}
    \vskip -2.0\baselineskip plus -1fil
\end{figure}

In embodied perception, robots observing dynamic environments from an egocentric perspective are increasingly expected to exhibit empathetic intelligence~\cite{duan2021survey,lin2026physbrain}. For applications such as home companionship and social interaction, the ability to understand emotional cues from visual context and respond appropriately is fundamental to natural and harmonious human–robot coexistence~\cite{mathur2024advancing,demelo2023social}. 
Recent advances in Multimodal Large Language Models (MLLMs) provide a strong technical foundation for this capability. By aligning vision and language representations, MLLMs enable cross-modal reasoning and complex video understanding~\cite{radford2021learning,alayrac2022flamingo,cheng2024videollama2,xu2025qwen25omni}, opening new avenues for embodied agents to interpret rich physical contexts and generate emotionally appropriate behaviors.

{To ground this capability, we focus on the compelling scenario of ``robotic movie companionship,'' where a robot must continuously interpret cinematic content and generate empathic responses aligned with the semantic narrative~\cite{paiva2017empathy}. While extensive research has addressed related sub-fields, existing paradigms fall short of the requirements for authentic embodied interaction. One line of work focuses on cinematic video understanding (\textit{e.g.}, narrative progression and event reasoning) using pristine movie clips~\cite{rao2024scriptviz,huang2020movienet,ataallah2025infinibench,rawal2024cinepile}. Another line emphasizes objective emotion recognition—identifying the affective states of characters depicted on screen from either third- or first-person views~\cite{lin2024e3,yang2023emoset,li2025ugotme,kosti2017emotic}. Concurrently, datasets in egocentric vision predominantly target action recognition and behavioral forecasting, largely overlooking affective modeling~\cite{cheng2024videgothink,grauman2022ego4d,damen2018epickitchens}.}

{Critically, deploying emotion understanding in physical human-robot interactions introduces severe domain and task discrepancies. First, regarding data distribution, an embodied agent typically observes media through a physical screen from an egocentric perspective. This visual input is inherently degraded by viewpoint shifts, viewing distance variations, screen glare, and environmental interference~\cite{martinmartin2021jrdb,sigurdsson2018actor}, presenting a stark contrast to the perfect digital frames used in conventional benchmarks. Second, and more fundamentally, regarding task formulation, the objective of an empathic agent is not merely to passively recognize the emotions of on-screen characters, but to actively generate its own subjective affective response conditioned on the observed content. Consequently, existing datasets and paradigms are misaligned with embodied viewing tasks and cannot directly support authentic emotion modeling for realistic, egocentric movie-watching scenarios.}

To address this limitation, we introduce \textit{EgoScreen-Emotion (ESE)}, a dataset specifically designed for emotion understanding under egocentric screen-view movie-watching conditions. \textit{ESE} is constructed through physically recorded egocentric screen-view captures in real-world environments, with systematic control over camera height, viewing distance, and illumination, thereby simulating the perceptual input of embodied agents in realistic companionship scenarios. Unlike conventional movie emotion datasets that treat emotion as an attribute of the video content itself, \textit{ESE} defines emotion as the affective response of the viewing agent to the observed scene and employs multi-rater annotations to ensure consistency and reliability.

{Building upon \textit{ESE}, we further develop a multimodal fine-tuning framework tailored to the perceptual characteristics of embodied viewing. To effectively balance narrative completeness and computational efficiency, we propose a memory-inspired hierarchical context modeling strategy. Instead of directly processing unbounded visual histories, our framework compresses long-term visual observations into structured textual abstractions, drastically reducing the visual token burden. This long-term narrative background is then dynamically integrated with short-term temporally sampled visual frames and synchronized audio windows. Furthermore, by introducing supervision with explicit reasoning rationales, we encourage the model to learn structured affective inference patterns. {Through comprehensive evaluations, we demonstrate that this multimodal integration and structured supervision substantially improves emotion understanding performance. Compared with a single-frame baseline (1F), incorporating multi-frame context, audio cues, and narrative summaries improves Accuracy from 57.66 to 63.01 and Macro-F1 from 18.95 to 25.95, demonstrating the effectiveness of structured multimodal context for embodied affective reasoning.}

Our egocentric embodied movie emotion understanding framework is illustrated in Fig.~\ref{fig:overview}.}

{In summary, our contributions are as follows:
\begin{compactitem}
    \item We introduce \textit{EgoScreen-Emotion (ESE)}, the first benchmark multimodal dataset for egocentric screen-view movie emotion understanding. Captured under realistic embodied viewing conditions and annotated by multiple raters, \textit{ESE} provides reliable supervision for modeling the affective responses of viewing agents.
    \item We propose a novel multimodal affective reasoning framework tailored for embodied movie-watching. Our approach uniquely integrates multi-frame visuals, semantic narratives, and audio signals, supervised by explicit reasoning rationales to learn structured inference patterns.

    \item We establish a comprehensive benchmark for egocentric screen-view movie emotion understanding. Extensive experiments and ablation studies show that our framework outperforms existing baselines in accuracy, stability, and interpretability under realistic egocentric screen-view conditions.
\end{compactitem}}

\vspace{-10pt}
\section{Related Work}

{\textbf{MLLMs in Embodied agents.}
Recent advancements in Multimodal Large Language Models (MLLMs), such as LLaVA~\cite{liu2024improved}, GPT-5~\cite{singh2025openai}, and Qwen-VL~\cite{wang2024qwen2}, have demonstrated unprecedented capabilities in integrating multimodal information with profound semantic reasoning. By aligning pre-trained vision encoders with powerful large language models, MLLMs excel at open-vocabulary visual understanding~\cite{mangalam2023egoschema}, spatial reasoning~\cite{zhou2024navgpt}, and complex visual question answering~\cite{xu2024lvlm}. Driven by these breakthroughs, a growing body of literature has begun integrating MLLMs into embodied applications to serve as the central cognitive core for robotic agents in fundamental tasks such as Vision-Language-Action (VLA) control~\cite{driess2023palm,zitkovich2023rt} and Vision-Language Navigation (VLN)~\cite{an2022bevbert}. Beyond acting as direct policy controllers, recent literature explores increasingly novel paradigms. For instance, MLLMs are now being leveraged as high-level semantic planners for long-horizon reasoning~\cite{huang2023voxposer} and spatial-aware cognitive cores integrated with 3D scene representations~\cite{gu2024conceptgraphs} to enable self-correction in complex physical environments.}

{\noindent \textbf{Multimodal Emotion Understanding.}
Emotion understanding plays a vital role in applications such as human-computer interaction, educational assistance, and psychological counseling~\cite{imani2019survey,mathur2024advancing,zadeh2018multimodal}. While early research primarily focused on single modalities~\cite{jiang2020dfew,lei2023instructerc,fan2021lssed}, recent works have shifted towards complex affective reasoning using multimodal data. Emotion-LLaMA~\cite{cheng2024emotion} integrates audio, visual, and textual inputs for multimodal emotion recognition and reasoning. HumanOmni~\cite{zhao2025humanomni} and OmniEmotion~\cite{yang2025omni} scale training with large video corpora to capture omni-modal interactions. The recently proposed Emotion-LLaMAv2~\cite{peng2026emotion} further introduces a large-scale unified multimodal dataset and proposes a model that supports structured affective perception within a unified architecture. However, these approaches or datasets~\cite{kossaifi2019sewa,kosti2017emotic,yang2023emoset,lin2024e3} predominantly process pristine digital videos and focus on objectively recognizing the emotions of human subjects, leaving a critical void in modeling how embodied agents themselves perceive and react to visual stimuli in physical environments.}

{As a result, existing literature lacks a naturalistic embodied setup where MLLMs can genuinely perceive and generate their own emotional responses to observed videos. We bridge this divide by introducing \emph{EgoScreen-Emotion}, pioneering the transition to subjective agent empathy. Concurrently, we propose a multimodal fine-tuning framework equipped with structured reasoning supervision, enabling accurate and stable emotion feedback generation under egocentric physical viewing conditions.}

\begin{table}[t]
\caption{
\textbf{Comparison across cinematic, emotion, and egocentric datasets.}
Emotion is divided into viewer-level and character-level modeling.
Conf. and Rat. denote confidence scores and textual rationales.
\#Anno. indicates the annotator number.
Emb. (Embodied) indicates whether the dataset is designed for embodied perception or robot-centric interaction scenarios.
}
\label{tab:dataset_compare}

\centering
\scriptsize
\setlength{\tabcolsep}{2pt}
\renewcommand{\arraystretch}{0.90}

\vskip -1.0\baselineskip plus -1fil
\resizebox{1.0\linewidth}{!}{
\begin{tabular}{@{}lcccccccccc@{}}
\toprule
\textbf{Dataset}
& \textbf{Domain}
& \textbf{Ego}
& \textbf{Emb.}
& \multicolumn{2}{c}{\textbf{Emotion}}
& \textbf{Modality}
& \multicolumn{2}{c}{\textbf{Annot.}}
& \textbf{\#Anno.}
& \textbf{\makecell{\#Annotated\\Units}} \\
\cmidrule(lr){5-6}\cmidrule(lr){8-9}
&  &  &  & \textbf{Viewer} & \textbf{Char.}
&  & \textbf{Conf.} & \textbf{Rat.}
&  &  \\
\midrule

AFEW~\cite{kossaifi2017afewva}
& movie & \ding{55} & \ding{55} & \ding{55} & \ding{51} & V,A & \ding{55} & \ding{55} & 1 & 30,000 \\

MovieNet~\cite{huang2020movienet}
& movie & \ding{55} & \ding{55} & \ding{55} & \ding{55} & V,A,T & \ding{55} & \ding{55} & -- & -- \\

InfiniBench~\cite{ataallah2024infinibench}
& movie & \ding{55} & \ding{55} & \ding{55} & \ding{55} & V,T & \ding{55} & \ding{55} & -- & 87,700 \\

CineTechBench~\cite{wang2025cinetechbench}
& movie & \ding{55} & \ding{55} & \ding{55} & \ding{55} & V & \ding{55} & \ding{55} & -- & 720 \\

MELD~\cite{poria2018meld}
& movie & \ding{55} & \ding{55} & \ding{55} & \ding{51} & V,A,T & \ding{55} & \ding{55} & 3 & 13,708 \\

\midrule

SEWA~\cite{kossaifi2019sewa}
& adverts & \ding{55} & \ding{55} & \ding{55} & \ding{51} & V,A & \ding{55} & \ding{55} & 5 & 1,990 \\

EMOTIC~\cite{kosti2017emotic}
& diverse & \ding{55} & \ding{55} & \ding{55} & \ding{51} & V & \ding{55} & \ding{55} & 3 & 23,788 \\

EmoSet~\cite{yang2023emoset}
& social & \ding{55} & \ding{55} & \ding{55} & \ding{51} & V & \ding{55} & \ding{55} & 10 & 118,102 \\

E3~\cite{lin2024e3}
& diverse & \ding{51} & \ding{51} & \ding{55} & \ding{51} & V,A,T & \ding{55} & \ding{51} & 3 & 81,248 \\

\midrule

VidEgoThink~\cite{cheng2024videgothink}
& daily & \ding{51} & \ding{55} & \ding{55} & \ding{55} & V,T & \ding{55} & \ding{55} & 1 & 4,993 \\

Ego4D~\cite{grauman2022ego4d}
& diverse & \ding{51} & \ding{55} & \ding{55} & \ding{55} & V,A,T & \ding{55} & \ding{55} & -- & 74,000 \\

EgoLife~\cite{yang2025egolife}
& daily & \ding{51} & \ding{55} & \ding{55} & \ding{55} & V,A,T & \ding{55} & \ding{55} & 1 & 3,000 \\

EPIC-KITCHENS~\cite{damen2018epic}
& kitchen & \ding{51} & \ding{55} & \ding{55} & \ding{55} & V,A,T & \ding{55} & \ding{55} & -- & 39,564 \\

\midrule

\textbf{ESE (Ours)}
& \textbf{movie}
& \textbf{\ding{51}}
& \textbf{\ding{51}}
& \textbf{\ding{51}}
& \textbf{\ding{55}}
& \textbf{V,A,T}
& \textbf{\ding{51}}
& \textbf{\ding{51}}
& \textbf{5}
& \textbf{28,667} \\

\bottomrule
\end{tabular}
}

\vskip -2.0\baselineskip plus -1fil
\end{table}

\vspace{-10pt}
\section{\textit{EgoScreen-Emotion} Dataset}
{While extensive multimodal and affective datasets exist across several domains, they predominantly suffer from two critical limitations: they either rely on non-egocentric viewpoints~\cite{kossaifi2017afewva,kossaifi2019sewa}, or they focus entirely on recognizing the objective emotions of third-party characters~\cite{lin2024e3}. Consequently, there is a complete absence of datasets dedicated to the subjective, intrinsic emotional responses of an agent from an egocentric perspective. To bridge this critical gap, we introduce \textit{EgoScreen-Emotion (ESE)}, comprising $224$ movie trailer clips and $28,667$ keyframes for frame-level emotion analysis. Each frame is annotated by five independent raters with multi-label confidence scores, yielding over $143,000$ human emotion annotations, with $10\%$ of samples additionally including textual rationales for interpretability. A detailed comparison with existing datasets is provided in Tab.~\ref{tab:dataset_compare}.}

\begin{figure}[t]
    \centering
    \includegraphics[width=\linewidth]{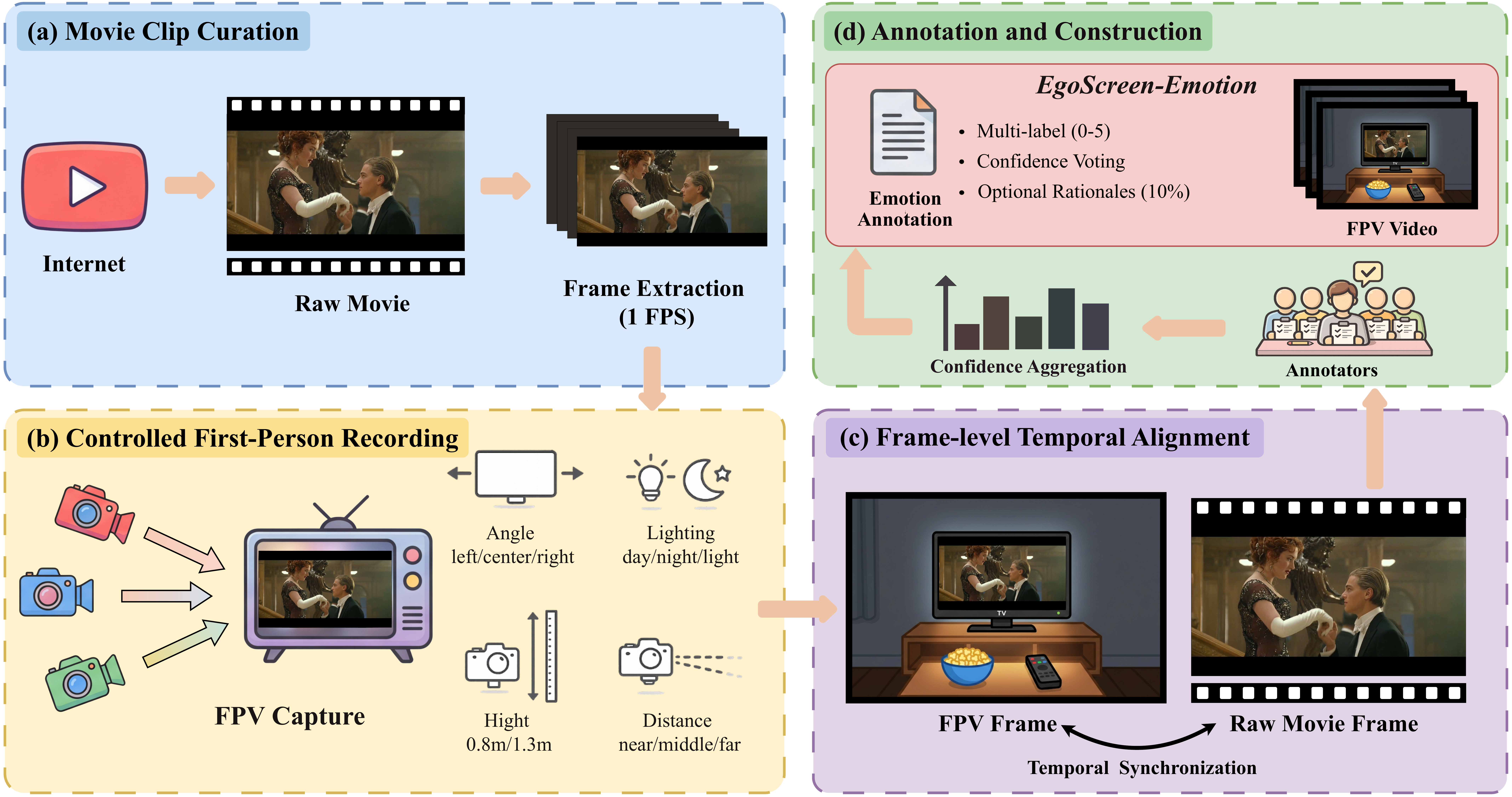}
    \vskip -0.5\baselineskip plus -1fil
    \caption{
    \textbf{\textit{EgoScreen-Emotion} dataset construction pipeline.}
    (a) Movie clip curation and frame extraction from raw movies.
    (b) Controlled first-person recording under simulated viewing conditions.
    (c) Frame-level temporal alignment between FPV and raw movie frames.
    (d) Emotion annotation with confidence aggregation.
    }
    \vskip -1.0\baselineskip plus -1fil
    \label{fig:construction_pipeline}
\end{figure}

\subsection{Data Collection and Preprocessing}
{The construction and preprocessing of \textit{ESE} dataset contains the following parts: movie clip curation, controlled first-person recording, temporal alignment, and frame-level sampling with annotation file generation, as illustrated in Fig.~\ref{fig:construction_pipeline}.}

\begin{figure}[t]
\centering


\begin{overpic}[width=0.38\linewidth]{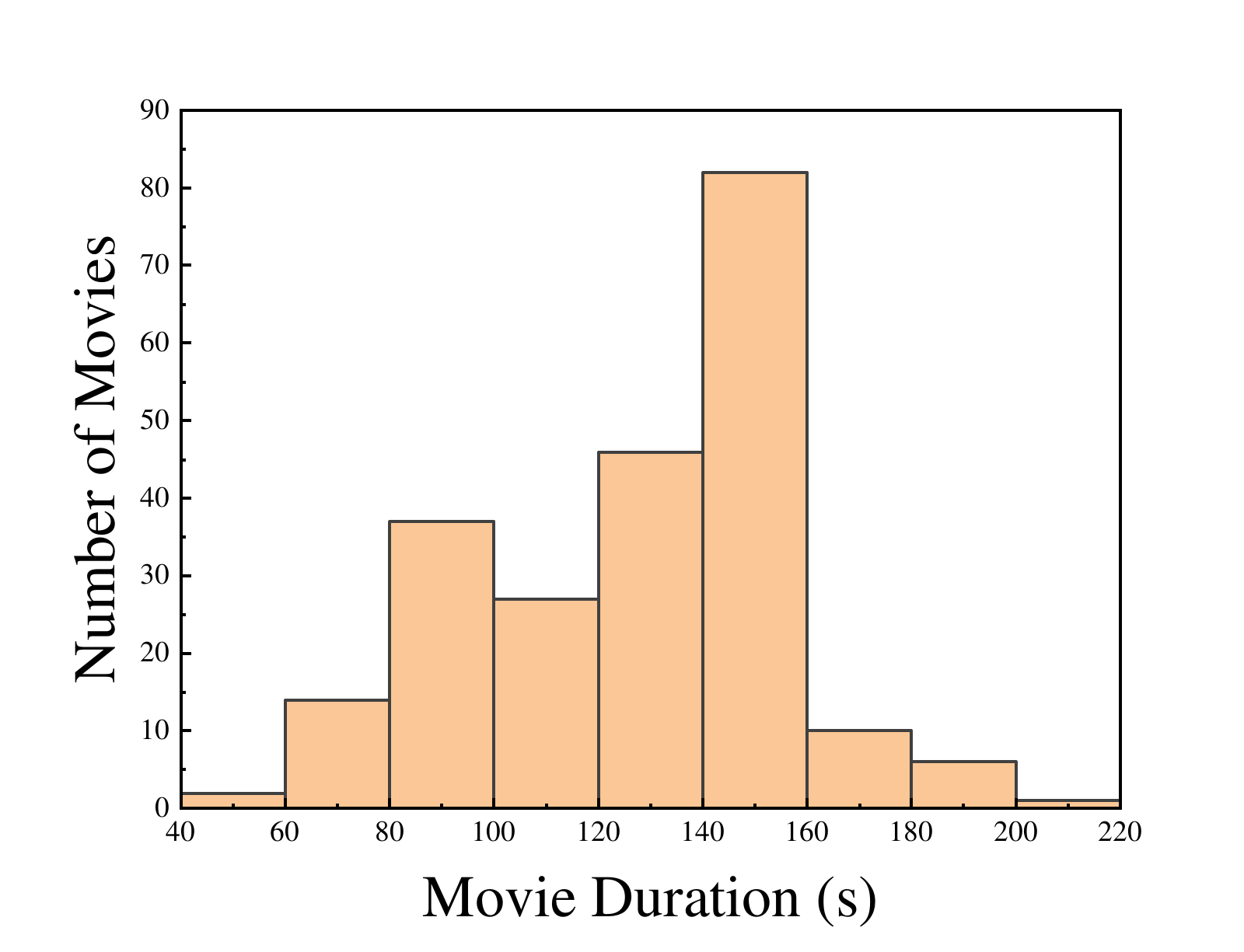}
\put(0,70){\small (a)}
\end{overpic}
\hspace{0.02\linewidth}
\begin{overpic}[width=0.38\linewidth]{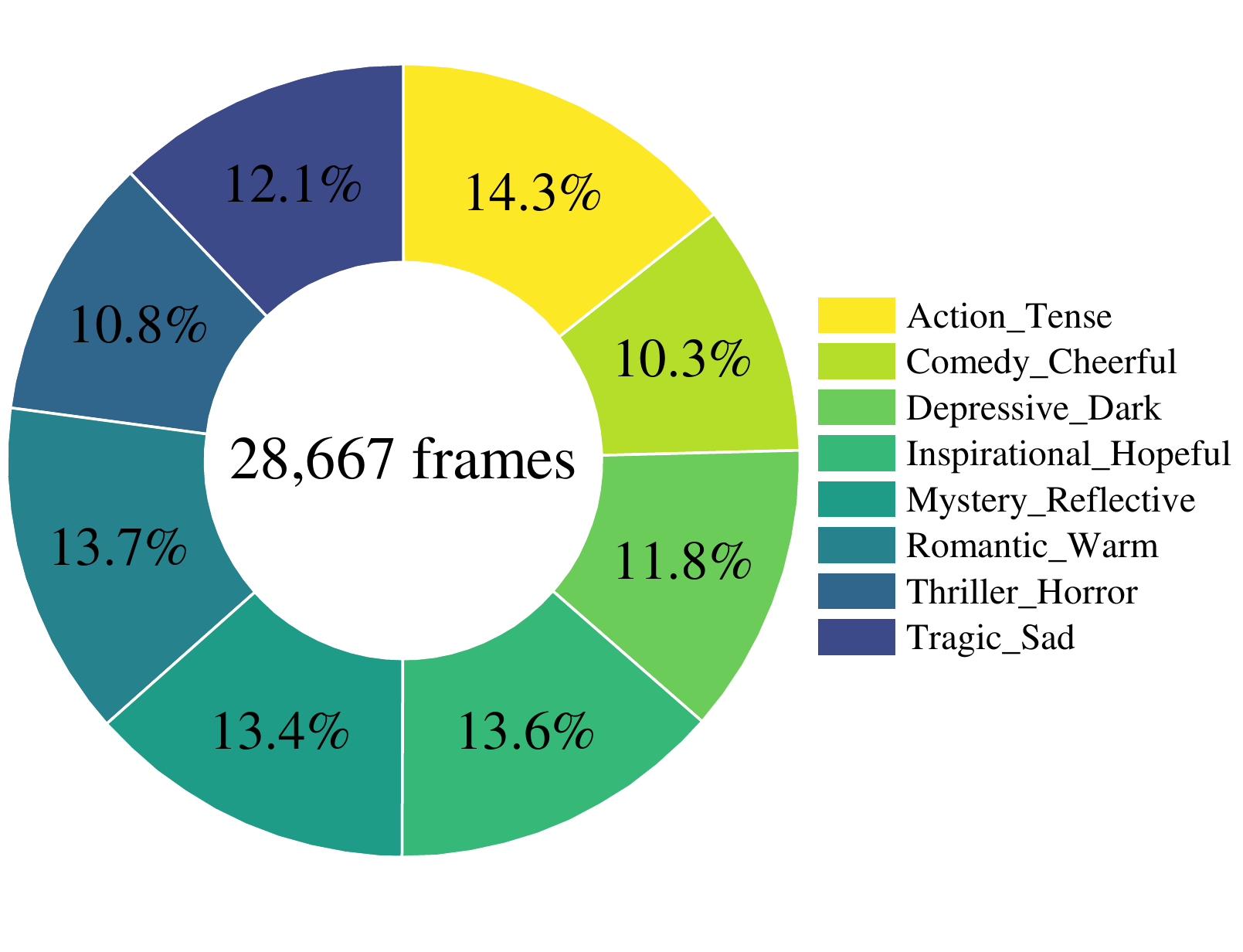}
\put(0,70){\small (b)}
\end{overpic}
\vskip -1.0\baselineskip plus -1fil
\caption{\textbf{Movie trailer statistics in \textit{EgoScreen-Emotion (ESE)}.}
(a) Trailer duration distribution.
(b) Distribution of movie clips across emotion categories.}

\label{fig:movie_stats}
\vspace{-3mm}
\vskip -1.0\baselineskip plus -1fil

\end{figure}

\noindent \textbf{Movie Clip Curation.}
{To ensure comprehensive content and structured affective coverage, \textit{ESE} collects $224$ publicly available movie trailers from YouTube, spanning $8$ representative emotion categories, as illustrated in Fig.~\ref{fig:movie_stats}.}
Each category contains a comparable number of films, providing balanced representation across emotional types. 
To ensure cinematic quality and narrative richness, the selected trailers are primarily drawn from Academy Award–winning or nominated films. 
Trailers are selected instead of full-length movies due to their condensed narrative structure and intensified emotional progression, which typically exhibit richer emotional transitions and broader affective diversity within shorter durations. 
This design facilitates fine-grained frame-level annotation while maintaining manageable temporal length.

\noindent \textbf{Controlled First-Person Recording.} 
{Embodied viewers operate in real-world environments and typically perceive movie content on a display from first-person view (FPV) at a natural viewing distance. Consequently, raw movie trailers alone are insufficient to capture the perceptual characteristics of embodied observation.}
To model embodied egocentric emotion understanding, each trailer is played on a display screen and recorded using a physically situated egocentric camera. The viewing configuration is systematically varied along multiple factors, including camera height ($0.8$m and $1.3$m), distance (near, middle, far), horizontal angle (left, center, right), and lighting condition (day, night, light on, light off). The selected camera heights are designed to approximate typical perception levels of embodied robots in companionship scenarios, corresponding to seated and standing viewpoints (e.g., similar to the height range of humanoid robots such as Unitree G1). 
This setup introduces realistic viewpoint distortion, perspective shift, and illumination variation compared to raw cinematic footage, forming a distinct egocentric screen-view visual domain.

\noindent \textbf{Temporal Alignment and Frame Sampling.} 
The recorded FPV videos are then temporally synchronized with the original movie streams at the frame level to ensure accurate correspondence. After alignment, frames are uniformly sampled at 1 FPS to balance temporal coverage and annotation feasibility. This process results in 28,667 aligned frames across all clips. 

In this procedure, the data are further processed via cropping and quality filtering to remove transitional or erroneous frames.

\noindent \textbf{Annotation File Construction.}
For each sampled frame, we generate a structured JSON annotation template containing predefined emotion categories and confidence fields. The finalized annotation files store multi-label emotion scores, annotator confidence levels, and optional textual rationales, forming the complete \textit{ESE} dataset.
{The annotation construction procedure are detailed in Sec.~\ref{sec: annotation construction}.}

\vspace{-10pt}
\subsection{Annotation Construction}
\label{sec: annotation construction}
As illustrated in Fig.~\ref{fig:construction_pipeline}(d), to systematically capture affective perception in first-person movie watching, we adopt a versioned, structured annotation schema and store all annotations in a unified JSON format. Each JSON file explicitly records the emotion class set, the confidence scale definition, video metadata (e.g., source video name and sampling rate), and per-frame annotation entries, ensuring traceability and extensibility. We support two complementary modes: \emph{simple} (confidence-only) and \emph{rationale} (confidence and overall textual justification). Both modes share the same emotion taxonomy with $10$ categories~\cite{ekman1992argument,kosti2017emotic,koelstra2012deap} ({including \texttt{angry}, \texttt{funny}, \texttt{fear}, \texttt{happy}, \texttt{sad}, \texttt{surprised}, \texttt{neutral}, \texttt{excited}, \texttt{touched} and \texttt{tense}}) and a discrete confidence scale. For each frame, annotators may select one or multiple emotions and assign a confidence score in \([0,5]\), where \(\,0\) denotes \emph{not selected} (\ie, no evidence for that emotion) and \(1\)–\(5\) indicate increasing certainty from ``most uncertain'' to ``most certain''. This multi-label confidence design preserves the inherent ambiguity and mixed affect commonly observed in cinematic scenes, providing richer supervision than rigid one-hot labels and enabling subsequent uncertainty-aware modeling.

\noindent\textbf{Annotation Protocol.}
{Each frame is independently annotated by $5$ human annotators following a standard guideline.}
Annotators are instructed to label the \emph{dominant emotion experienced as a viewer} when observing the current frame, rather than describing characters' facial expressions or objective scene semantics, to avoid conflating visual semantics with subjective affect. The interface pre-fills all emotion categories for every frame; annotators only need to assign scores to selected emotions, while unselected categories remain \(0\), resulting in sparse yet interpretable records. In \emph{rationale} mode, a subset of frames additionally includes a concise free-text justification that summarizes key visual cues (e.g., lighting, color tone, composition, posture, and atmosphere). These rationales are not used for label aggregation but serve as auxiliary signals for explainable and reasoning-oriented models.

\noindent\textbf{Confidence-Summed Voting.}
To derive a single evaluation label per frame, we perform confidence-summed voting across annotators. Let \(s_{i,c}\in[0,n]\) be annotator \(i\)'s score for emotion class \(c\) (with \(0\) for unselected). We compute the aggregated score for each class:
\begin{equation}
S_c=\sum_{i=1}^{n}s_{i,c}.
\end{equation}

{The final dominant emotion label is determined by the highest aggregated score:}
\begin{equation}
L=\arg\max_{c} S_c.
\end{equation}

{This aggregated approach systematically incorporates both \emph{support frequency} and \emph{support strength}, thereby conferring enhanced stability compared to rudimentary majority voting.
For instance, strong high-confidence votes originating from a minority contingent of annotators are not arbitrarily dismissed. 
We preserve the full per-frame confidence distributions and optional rationales, enabling in-depth uncertainty analysis and future research beyond single-label assessment.}

\begin{figure}[t]
    \centering
    
    \makebox[\linewidth][c]{
        
        \raisebox{-0.14cm}{
            \begin{overpic}[width=0.38\linewidth]{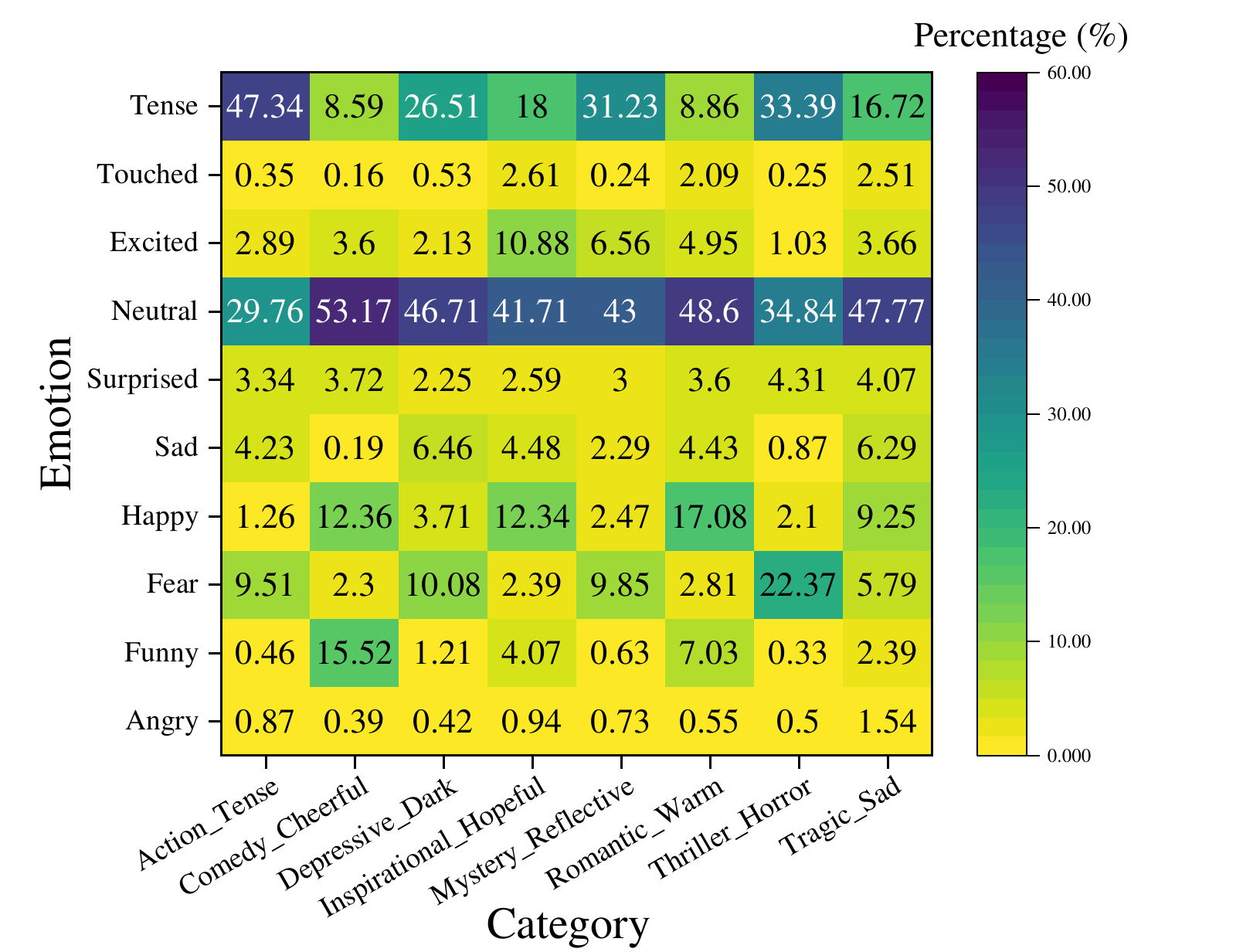}
                \put(0, 73){\textbf{(a)}} 
            \end{overpic}
        }
        
        \hspace{-0.06\linewidth} 
        
        \begin{overpic}[width=0.38\linewidth]{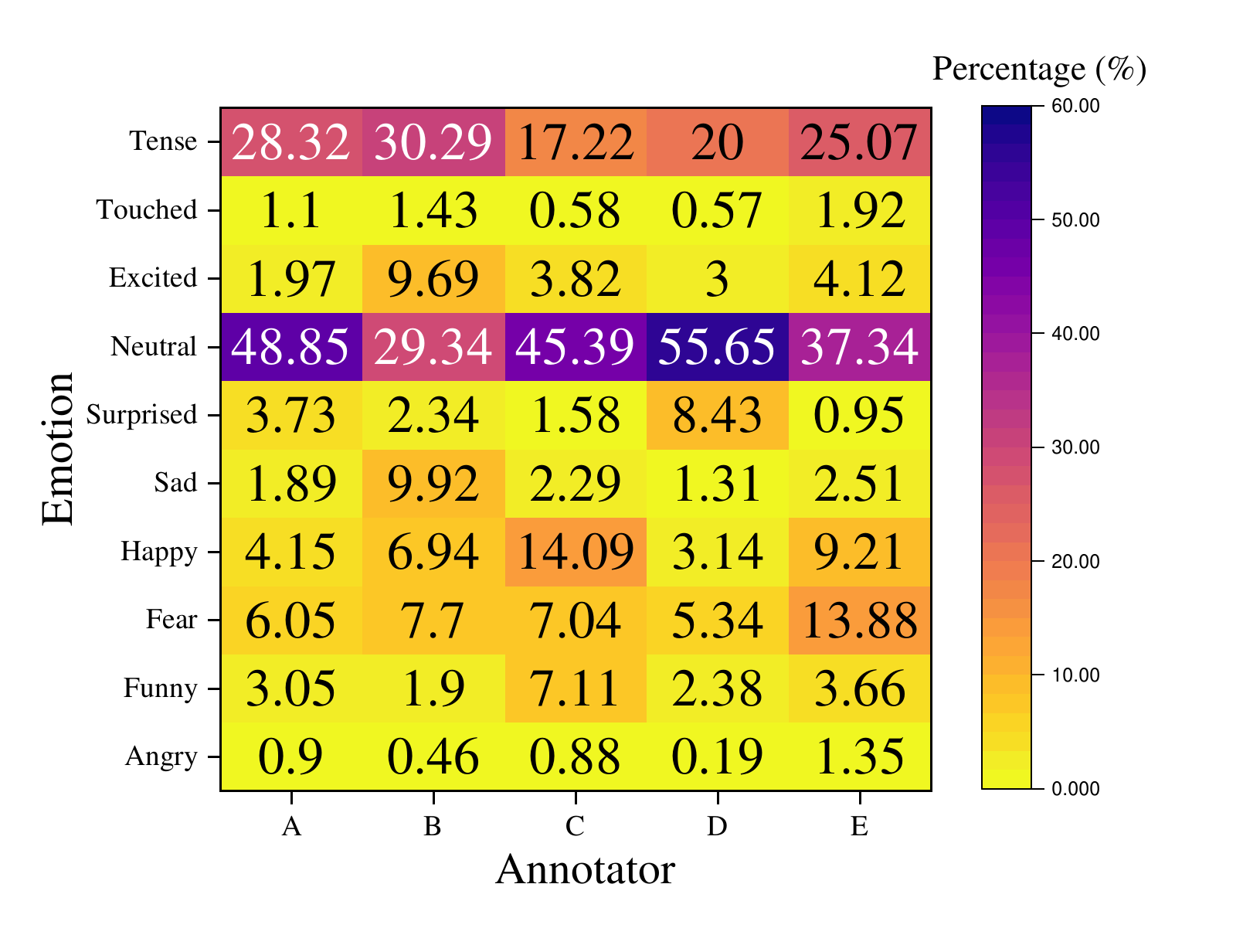}
            \put(2, 70){\textbf{(b)}}
        \end{overpic}
        
        \hspace{-0.05\linewidth} 
        
        \begin{overpic}[width=0.34\linewidth]{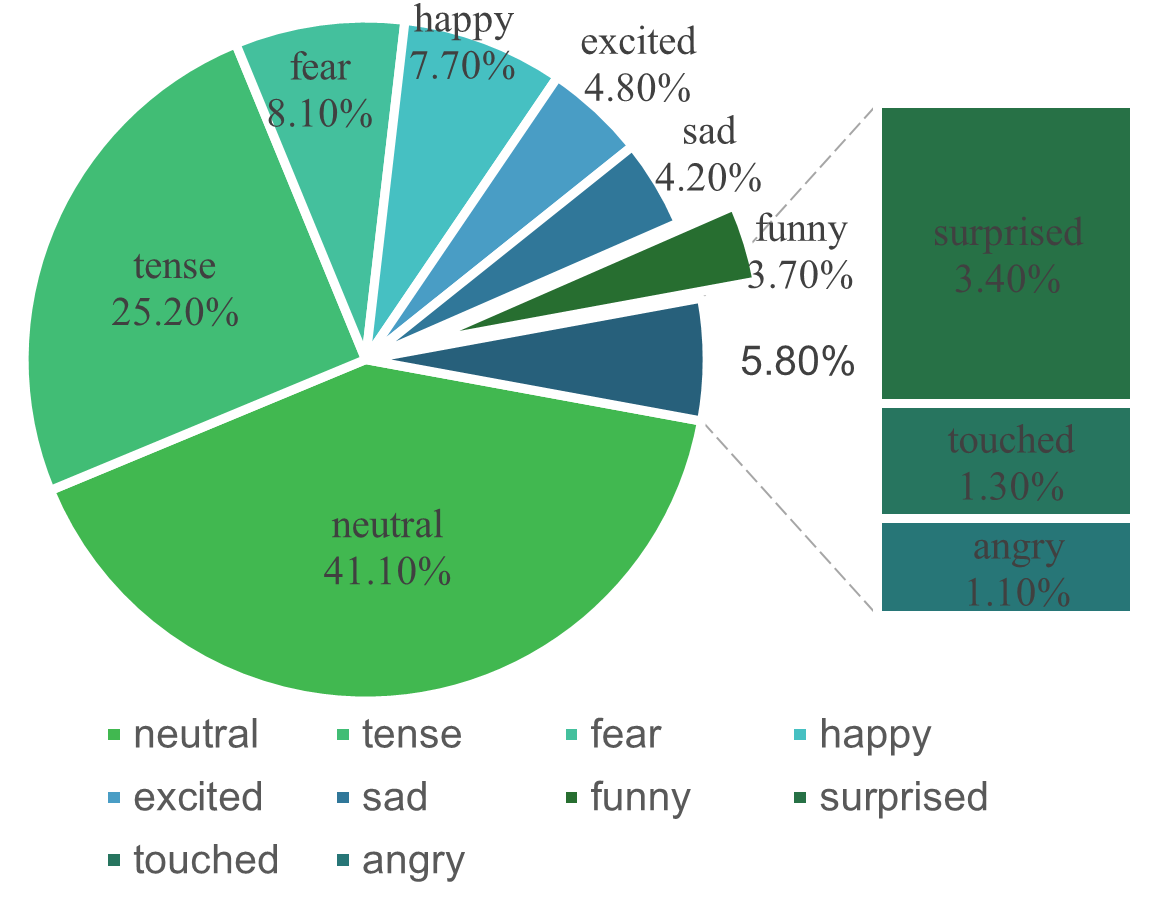}
            \put(2, 79){\textbf{(c)}}
        \end{overpic}
        
    } 
    
    \vspace{-2mm}
    \vskip -1ex
    \caption{
    \textbf{Statistical analysis of the \textit{EgoScreen-Emotion} dataset.}
    (a) Distribution of emotion categories across movie genres.
    (b) Emotion distribution annotated by different annotators.
    (c) Overall emotion category distribution in the dataset.
    }
    \label{fig:data_analysis}
    
\vskip -2.0\baselineskip plus -1fil
\end{figure}

\vspace{-10pt}
\subsection{Dataset Analysis}
{We conduct further statistical analysis on the \textit{EgoScreen-Emotion} dataset to ensure the reliability of its annotations, including emotion category distribution, multi-emotion ratio and annotator bias analysis.
The statistics regarding the emotion distributions across different cinematic genres, inter-annotator bias, and the overall emotion label distribution of the ESE dataset are visualized in Fig~\ref{fig:data_analysis}.}

\noindent \textbf{Emotion Category Distribution.}
The overall statistics exhibit a naturally long-tailed distribution consistent with cinematic narrative structure. The \texttt{neut- ral} category accounts for the largest proportion (over $40\%$), followed by \texttt{tense} and \texttt{fear}. This pattern aligns with the prevalence of transitional and context-setting shots in movie trailers. 
Across movie categories (\textit{e.g.}, Action, Comedy, Depressive, Romantic), dominant emotion categories vary systematically. For instance, in Action subsets, the proportions of \texttt{tense} and \texttt{fear} increase significantly, whereas in Romantic and Inspirational subsets, \texttt{happy} and \texttt{excited} appear more frequently. This consistency between thematic intent and annotated emotion distribution suggests that the labels capture semantic affective tendencies rather than random assignment.

\noindent \textbf{Multi-Emotion Ratio.}
Our analysis of label density reveals that over 80\% of frames contain a single dominant emotion ($k=1$), approximately 15\%--20\% include two emotion labels ($k=2$), and fewer than 1\% involve three or more labels. These statistics indicate that most frames convey relatively clear emotional signals, while a meaningful subset reflects mixed or transitional emotional states. The multi-label confidence mechanism therefore preserves nuanced cinematic emotions without enforcing artificial single-label constraints.

\noindent \textbf{Annotator Bias Analysis.}
To examine potential systematic personal bias, we compute emotion selection proportions for each annotator. All annotators assign a relatively high percentage of \texttt{neutral} labels, and their overall distribution structures remain consistent. Minor differences exist in specific categories (\textit{e.g.}, \texttt{happy}, \texttt{tense}, \texttt{fear}), but no annotator shows abnormal dominance in a single category (\textit{e.g.}, exceeding 70\%). These results suggest normal individual variability without systematic bias, indicating good group-level stability.

\vspace{-10pt}
\subsection{Privacy Protection}
The dataset is constructed from publicly available movie trailers collected from YouTube and is released for academic research purposes only. The released data consist of processed egocentric recordings (video, audio, frames, and annotations) captured during controlled viewing conditions, rather than the original movie trailer files. No real viewers are recorded and no personally identifiable information (PII) is collected.

\begin{center}
\fcolorbox{black}{gray!10}{
\parbox{0.95\linewidth}{
\textbf{Dataset Contribution.}
We introduce \textbf{\textit{EgoScreen-Emotion (ESE)}}, the first cross-domain visual dataset for embodied emotion understanding in movie-watching scenarios.
ESE captures movie content from a realistic \textit{egocentric screen-view} setting and provides viewer-level emotion annotations.
}}
\end{center}

\vspace{-10pt}
\section{Baseline Modeling Framework}

\begin{figure}[t]
    \centering
    \includegraphics[width=0.9\linewidth]{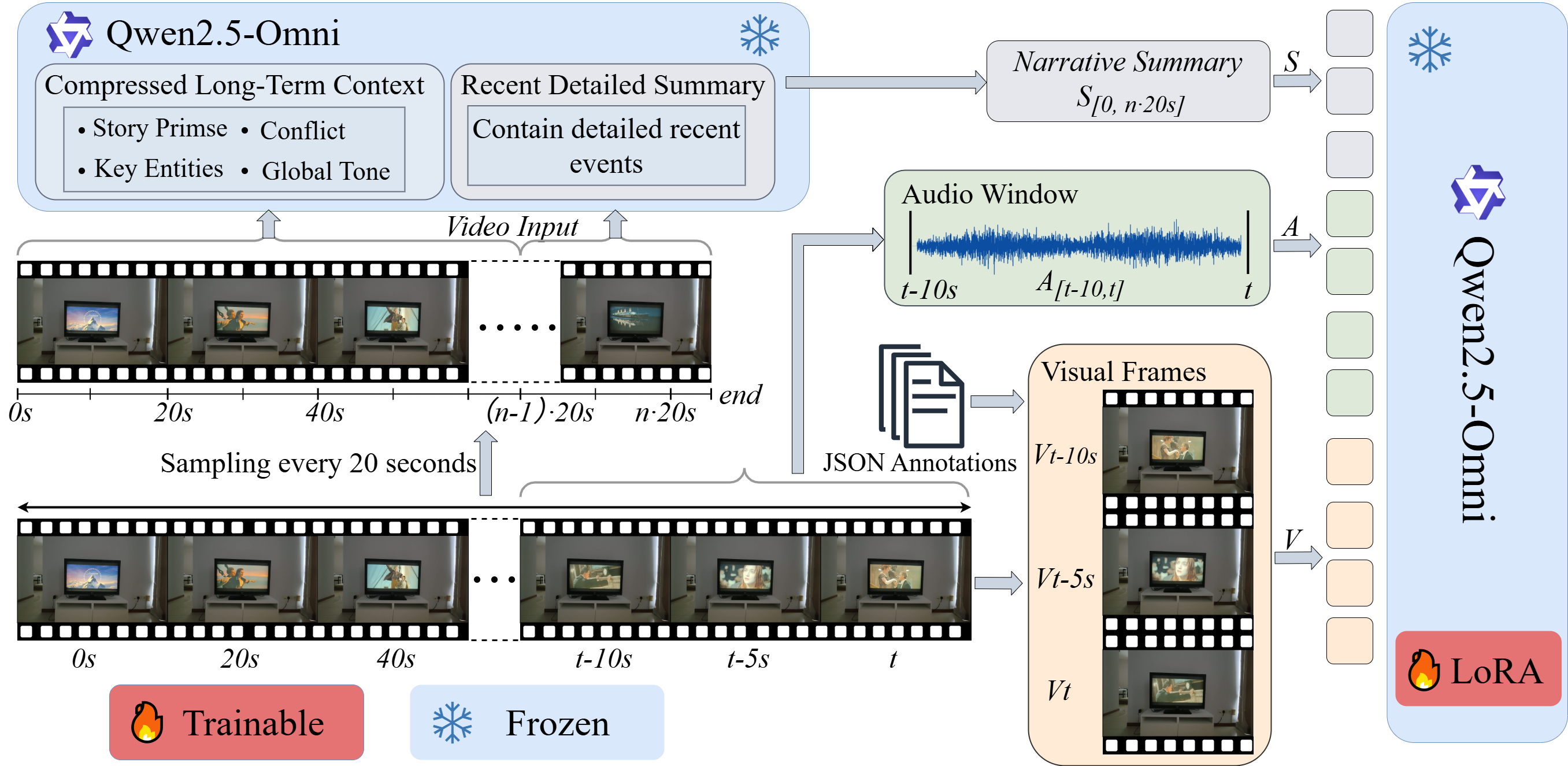}
    \vskip -2ex
    \caption{
\textbf{Baseline framework for egocentric screen-view movie emotion understanding.}
At each timestamp $t$, we predict the viewer-oriented emotion $\hat{y}_t$ by conditioning Qwen2.5-Omni on a multimodal triplet $(\mathbf{V}_t,\mathbf{A}_t,\mathbf{S}_t)$: {visual keyframes} $\mathbf{V}_t$ sampled at 5-second intervals (up to three frames at $t$, $t\!-\!5$s, $t\!-\!10$s), an {audio window} $\mathbf{A}_t$ covering the previous 10 seconds, and a {memory-inspired narrative context} $\mathbf{S}_t$ that combines a recent detailed 20-second segment summary with a compressed long-term background (premise, entities, conflict, global tone) produced every 20 seconds and further abstracted over time. 
The backbone is kept frozen and adapted to the ESE domain via LoRA fine-tuning, enabling long-context affective reasoning under realistic screen-view observations.
}
    \vskip -4ex
    \label{fig:sft_framework}
\end{figure}

In embodied movie companionship scenarios, emotion responses are not only determined by instantaneous perceptual stimuli, but are also shaped by the accumulated narrative context. Instead of classifying the emotions of on-screen characters, we aim to generate the viewing agent's intrinsic affective response under egocentric conditions.
We formulate this task as a multimodal conditional prediction problem:
\begin{equation}
\hat{y}_t = f_\theta(\mathbf{V}_t, \mathbf{A}_t, \mathbf{S}_t),
\end{equation}
where $\hat{y}_t$ denotes the viewer-oriented emotion at time $t$, 
$\mathbf{V}_t$ represents visual keyframes, 
$\mathbf{A}_t$ denotes an audio window, and 
$\mathbf{S}_t$ encodes structured narrative context. 
The function $f_\theta(\cdot)$ is instantiated by Qwen2.5-Omni-7B~\cite{xu2025qwen25omni} and adapted to the egocentric movie-viewing domain via LoRA-based fine-tuning on the \textit{ESE} dataset. Fig.~\ref{fig:sft_framework} illustrates the overall baseline modeling framework.

\noindent \textbf{Short-Term Perceptual Modeling (V and A).}
To capture local emotional triggers, we construct temporally aligned visual and auditory inputs.
Visual observations are sampled at 5-second intervals. 
Let $t$ denote the current timestamp. The visual input is defined as
\begin{equation}
\mathbf{V}_t = \{ I_{t-\ell} \mid \ell \in \mathcal{L}_t \},
\end{equation}
where $\mathcal{L}_t$ expands progressively with time: 
for $t<5$, only the current frame is used; 
for $5 \le t < 10$, the current frame and the frame 5 seconds earlier are used; 
for $t \ge 10$, three frames are included: the current frame, the frame 5 seconds earlier, and the frame 10 seconds earlier. 
This design captures short-term temporal dynamics while maintaining a bounded visual token budget.

The audio input is defined as a sliding window covering at most the previous 10 seconds:
\begin{equation}
\mathbf{A}_t = A[\max(0, t-10),\, t],
\end{equation}
which provides complementary affective cues such as speech tone, dialogue intensity, and background music. 
When $t<10$, the window is truncated accordingly.

\noindent \textbf{Long-Term Narrative Modeling (S).}
Emotional responses during movie watching are influenced by long-term narrative progression. 
Inspired by human memory patterns---where recent events are remembered in detail while earlier events are abstracted into higher-level summaries---we adopt a two-level narrative representation consisting of a recent detailed summary and a compressed long-term background.
We divide each video into 20-second segments and generate segment-level summaries automatically using Qwen2.5-Omni-7B given the video input. 
Let
\begin{equation}
k = \left\lfloor \frac{t}{20} \right\rfloor
\end{equation}
denote the number of completed 20-second segments at time $t$. 
If $t<20$, no summary is available. 
If $t\ge 20$, the most recently completed segment is treated as the recent detailed summary. 
If $t\ge 40$, all earlier segments are further compressed---again using the same backbone model---into a long-term background abstraction capturing story premise, key entities, conflict relations, and global narrative tone.

The final textual context is constructed as:
\begin{equation}
\mathbf{S}_t = \Phi(\mathbf{S}^{\text{long}}_t,\, \mathbf{S}^{\text{recent}}_t),
\end{equation}
where $\Phi(\cdot)$ denotes structured concatenation under a fixed prompt template.

\noindent \textbf{Unified Multimodal Input.}
The triplet $(\mathbf{V}_t, \mathbf{A}_t, \mathbf{S}_t)$ is jointly fed into Qwen2.5-Omni-7B, which is fine-tuned using LoRA~\cite{hu2022lora} adapters to generate viewer-conditioned emotional responses. 
This design offers two advantages. 
First, the hierarchical textual abstraction mirrors human narrative memory, preserving coherence while preventing context explosion. 
Second, directly replaying long visual histories would cause visual tokens to grow linearly with time, increasing computational cost and inference latency. 
By compressing historical context into token-efficient textual summaries, we achieve a practical balance between narrative completeness and computational efficiency for embodied movie-watching interaction.

\vspace{-10pt}
\section{Experiments}
\vspace{-10pt}
In this section, we evaluate the \textit{EgoScreen-Emotion} dataset and our proposed framework. We formulate emotion prediction as a 10-class classification task, deriving ground-truth labels via the confidence-summed aggregation detailed in Sec.~\ref{sec: annotation construction}. To rigorously account for the long-tailed distribution, we report Accuracy, Macro-F1, and Weighted-F1. Our experiments systematically investigate: (i) the cross-domain generalization between raw cinematic footage and first-person view (FPV) data to justify FPV-specific supervision (Sec.~\ref{sec:cross_domain}); (ii) the effectiveness of our structured temporal and multimodal modeling  for embodied movie-watching scenarios (Sec.~\ref{sec:structured_context}); and (iii) a comprehensive benchmark against strong open- and closed-source baselines (Sec.~\ref{sec:quantitative_comparison}).

\vspace{-10pt}
\subsection{Cross-Domain Generalization Analysis}
\label{sec:cross_domain}

\begin{table}[t]
\centering
\caption{
Overall performance across four training/testing settings, illustrating the domain gap between raw cinematic frames and first-person screen-view (FPV) observations.
}
\label{tab:four_runs_overall}
\small
\setlength{\tabcolsep}{6pt}
\begin{tabular}{@{}lccc@{}}
\toprule
\rowcolor{gray!20}Setting & Acc & Macro-F1 & Weighted-F1 \\
\midrule
FPV$\rightarrow$FPV   & 57.74 & 20.61 & 55.18 \\
Raw$\rightarrow$Raw   & 63.33 & 27.99 & 60.76 \\
Raw$\rightarrow$FPV   & 55.75 & 16.69 & 46.47 \\
FPV$\rightarrow$Raw   & 61.19 & 25.19 & 58.68 \\
\bottomrule
\end{tabular}
\end{table}

To assess whether FPV-specific supervision is required for egocentric movie emotion understanding, we perform controlled cross-domain experiments using Qwen3-VL-8B~\cite{bai2025qwen3vl} as a unified backbone. The model is fine-tuned under each training domain and evaluated on the designated test domain. We construct four training-testing combinations across Raw cinematic footage and physically recorded egocentric screen-view data: FPV$\rightarrow$FPV, Raw$\rightarrow$Raw, Raw$\rightarrow$FPV, and FPV$\rightarrow$Raw.

As shown in Table~\ref{tab:four_runs_overall}, the Raw$\rightarrow$Raw setting achieves strong in-domain performance. However, when evaluated on FPV data (Raw$\rightarrow$FPV), both Accuracy and Macro-F1 decrease markedly, with Macro-F1 dropping by over 40\% relative to the in-domain counterpart. This performance degradation indicates limited cross-domain transfer from clean cinematic footage to realistic first-person viewing conditions.
Conversely, the FPV$\rightarrow$FPV configuration substantially mitigates this drop, demonstrating the importance of domain-consistent supervision. A comparable discrepancy is observed in the FPV$\rightarrow$Raw setting, further confirming a systematic distribution shift between Raw and FPV domains.
These results collectively suggest that viewpoint variation, reflection artifacts, illumination changes, and environmental interference alter the visual statistics and contextual cues of movie frames, making Raw-only supervision insufficient for robust emotion understanding in embodied egocentric scenarios.

\begin{table}[t]
\centering
\caption{
Effect of rationale supervision under single-frame input.
Both settings use identical visual input (1F), 
while the second setting incorporates rationale-based supervision 
on 10\% of training samples.
}
\label{tab:ablation_rationale}
\small
\setlength{\tabcolsep}{6pt}
\begin{tabular}{lccc}
\toprule
\rowcolor{gray!20}Setting & Acc & Macro-F1 & Weighted-F1 \\
\midrule
1F & 57.66 & 18.95 & 53.65  \\
\makecell[l]{1F +\\ Rationale (10\%)} & 58.32 & 20.78 & 55.52 \\
\bottomrule
\end{tabular}
\end{table}

\vspace{-10pt}
\subsection{Structured Temporal and Multimodal Context Modeling}
\label{sec:structured_context}
In egocentric movie-watching scenarios, a viewer's affective response is rarely determined by an isolated frame.
Instead, emotions are shaped by \emph{short-term visual dynamics} (e.g., sudden motion or tension escalation), 
\emph{long-term narrative context} (e.g., prior events that explain the current conflict), 
and \emph{audio cues} (e.g., dialogue tone and background music).
Motivated by this observation, we study how structured temporal and multimodal context helps adapt a strong multimodal backbone, Qwen2.5-Omni-7B~\cite{xu2025qwen25omni}, to realistic egocentric screen-view emotion understanding.

\begin{table}[t]
\centering
\caption{
Ablation study of structured temporal and multimodal context modeling 
based on Qwen2.5-Omni-7B.
All 3F-based configurations use three frames sampled at 5-second intervals.
}
\label{tab:ablation_structured}
\small
\setlength{\tabcolsep}{6pt}
\begin{tabular}{lccc}
\toprule
\rowcolor{gray!20}Setting & Acc & Macro-F1 & Weighted-F1 \\
\midrule
1F & 57.66 & 18.95 & 53.65 \\
3F & 61.06 & 23.42 & 57.17 \\
3F+Aud & 61.92 & 25.55 & 58.97 \\
3F+Aud+Narr & 63.01 & 25.95 & 60.70 \\
\bottomrule
\end{tabular}
\end{table}

\noindent \textbf{Rationale supervision as structured affective reasoning.}
We first isolate the effect of rationale-based supervision while keeping the visual input identical. As shown in Table~\ref{tab:ablation_rationale},both settings use the same single-frame visual input (1F), while the second additionally introduces rationale supervision on 10\% of the training samples. We observe consistent improvements across Accuracy, Macro-F1, and Weighted-F1, suggesting that rationale supervision improves prediction reliability. This lightweight supervision acts as a reasoning scaffold, encouraging the model to associate emotion predictions with explicit evidence rather than relying on superficial correlations.

\noindent \textbf{Ablating temporal and multimodal context.}
Next, we progressively evaluate the components of our structured context modeling framework (Table~\ref{tab:ablation_structured}).
Compared to 1F, using three frames sampled at 5-second intervals (3F) improves performance, suggesting that short-term temporal cues already provide useful signals for affective inference under egocentric recording noise.
Building upon this temporal context, we further introduce synchronized audio input (\texttt{3F + Aud}).
The additional improvement suggests that acoustic signals such as dialogue intensity, speech prosody, and background music provide complementary emotional evidence that cannot be captured by visual information alone.
Finally, incorporating narrative summaries (\texttt{3F + Aud + Narr}) leads to the best overall performance. This result highlights the importance of long-horizon narrative context, as many emotional reactions depend not only on immediate visual stimuli but also on accumulated story context that explains the current scene.

\begin{table}[t!]
\caption{
Comparison on the \textit{EgoScreen-Emotion} test set.
All baselines are evaluated in a zero-shot manner with a unified prompt,
while ours is fine-tuned on training set.
}
\vskip -2ex
\label{tab:main_comparison}
\centering
\small
\setlength{\tabcolsep}{6pt}
\resizebox{.8\textwidth}{!}{\begin{tabular}{@{}lcccc@{}}
\toprule
\rowcolor{gray!20}\textbf{Model} & \textbf{Params} & \textbf{Acc} & \textbf{Macro-F1} & \textbf{Weighted-F1} \\
\midrule
Qwen2.5-Omni~\cite{xu2025qwen25omni} & 7B & 55.22 & 16.36 & 46.76 \\
Qwen3VL~\cite{bai2025qwen3vl} & 8B & 49.00 & 20.71 & 50.26 \\
LLaVA-1.6~\cite{liu2024improved} & 13B & 52.55 & 8.75 & 40.38 \\
LLaVA-OneVision-1.5~\cite{an2025llavaonevision}   & 8B & 39.18 & 14.08 & 37.23 \\
InternVL2~\cite{chen2024internvl} & 8B & 27.47 & 11.92 & 34.38 \\
\midrule
Qwen3.5-plus & -- & 54.07 & 16.88 & 49.58 \\
GPT-5.2~\cite{singh2025openai} & -- & 56.87 & 18.36 & 52.85 \\
Gemini-2.5-flash~\cite{comanici2025gemini25} & -- & 52.62 & 23.50 & 53.52 \\
\midrule
\textbf{QwenSFT (Ours, 1F)} & \textbf{7B} & \textbf{57.66} & \textbf{18.95} & \textbf{53.65}  \\
\bottomrule
\end{tabular}}
\vskip -3ex
\end{table}

\vspace{-10pt}
\subsection{Quantitative Comparison with Strong Baselines}
\label{sec:quantitative_comparison}
\noindent\textbf{Quantitative comparison.}
We compare our model with representative open-source and closed-source multimodal models on the \textit{EgoScreen-Emotion} test set (Table~\ref{tab:main_comparison}). 
To ensure a fair comparison with existing models that perform emotion inference from a single visual input, we report the single-frame version of our model (1F) in this table.
All baselines are evaluated in a zero-shot setting using a unified prompt, while our model (\textbf{QwenSFT}) is fine-tuned on the ESE training set. 
Among open-source models, Qwen2.5-Omni~\cite{xu2025qwen25omni}, Qwen3VL~\cite{bai2025qwen3vl}, LLaVA-1.6~\cite{liu2024improved}, LLaVA-OneVision~\cite{an2025llavaonevision}, and InternVL2~\cite{chen2024internvl} show varying levels of performance, highlighting the difficulty of egocentric screen-view emotion understanding under a long-tailed label distribution. 
In contrast, our fine-tuned model achieves the best overall accuracy (57.66) and the highest Weighted-F1 (53.65) with a comparable parameter scale (7B). 
Compared with strong closed-source systems such as GPT-5.2~\cite{singh2025openai} and Gemini-2.5-flash~\cite{comanici2025gemini25}, our model achieves competitive performance, demonstrating the effectiveness of domain-specific fine-tuning on ESE.

\begin{figure}[t]
    \centering
    \includegraphics[width=\linewidth]{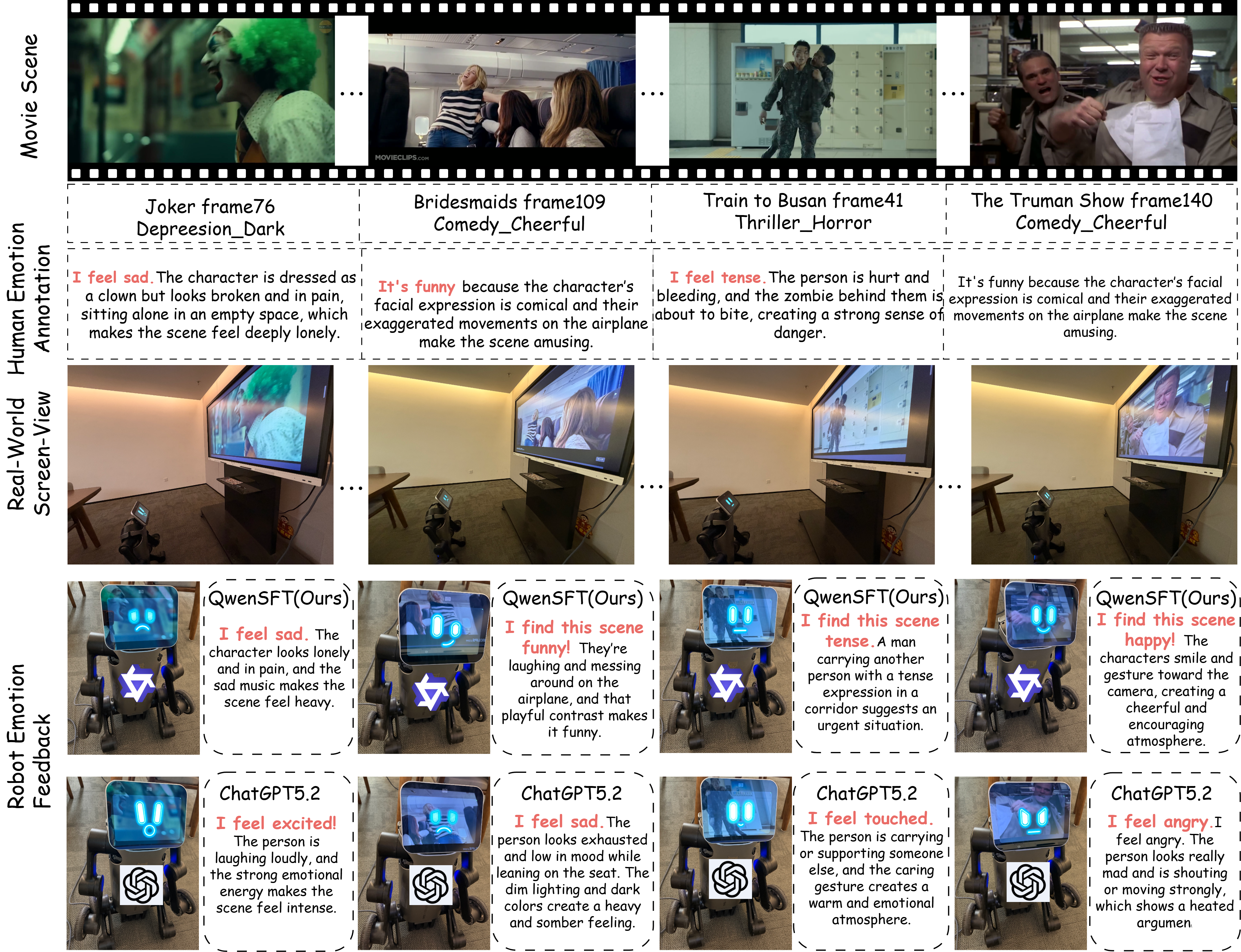}
    \vskip -0.5\baselineskip plus -1fil
    \vskip -1ex
    \caption{
    \textbf{Qualitative comparison under real-world egocentric screen-view conditions.}
    From top to bottom: (1) representative movie scenes,
    (2) human emotion annotations,
    (3) physically recorded egocentric screen-view observations,
    (4) emotion feedback generated by \textbf{QwenSFT (Ours)},
    and (5) emotion feedback generated by ChatGPT-5.2.
    All models are evaluated under identical real-world observations.
    }
    \label{fig:embodied_real}
    \vskip -5ex
\end{figure}

\noindent\textbf{Qualitative comparison in real-world screen-view scenarios.}
To further examine model behavior in realistic settings, we present a qualitative comparison under real-world egocentric screen-view conditions (Fig.~\ref{fig:embodied_real}). 
In this setup, a robot observes movie content from a physical screen and generates emotion feedback based on the perceived scene. 
Compared with responses generated by ChatGPT-5.2 under identical observations, our model produces emotion predictions that are more consistent with human annotations. 
These results indicate that the proposed framework can generate coherent viewer-oriented emotional responses in practical robotic movie companionship scenarios.

\vspace{-10pt}
\section{Conclusion and Future Work}
\vspace{-5pt}
In this work, we presented \textit{EgoScreen-Emotion (ESE)}, the first benchmark dataset for egocentric screen-view movie emotion understanding, shifting the focus from cinematic footage-based emotion analysis to viewer-centered affective understanding in embodied settings. 
To address the perceptual and computational challenges of realistic embodied movie-watching scenarios, we further proposed a memory-inspired multimodal framework. 
By compressing long-term visual histories into structured textual summaries and incorporating explicit reasoning supervision, our approach improves the accuracy, stability, and interpretability of emotion understanding. 
We hope that \textit{ESE} and the proposed framework will provide a useful benchmark and facilitate future research on emotion understanding under realistic egocentric perception for embodied agents.

\noindent \textbf{Future Work:} 
Building upon the current framework, future research could further incorporate viewer-centric signals, such as the audience’s facial expressions and vocal reactions during movie watching. 
Integrating these cues with screen-view observations may enable a more holistic understanding of both movie semantics and the surrounding affective context, ultimately supporting richer emotion interaction for embodied agents.

%
%
\clearpage
\bibliographystyle{splncs04}
\bibliography{main}

\clearpage
\appendix

\section*{Supplementary Material}

\section{Additional Dataset Statistics}
\begin{figure}[h]
\centering

\begin{overpic}[width=0.42\linewidth,trim=4cm 2cm 4cm 1.9cm,clip]{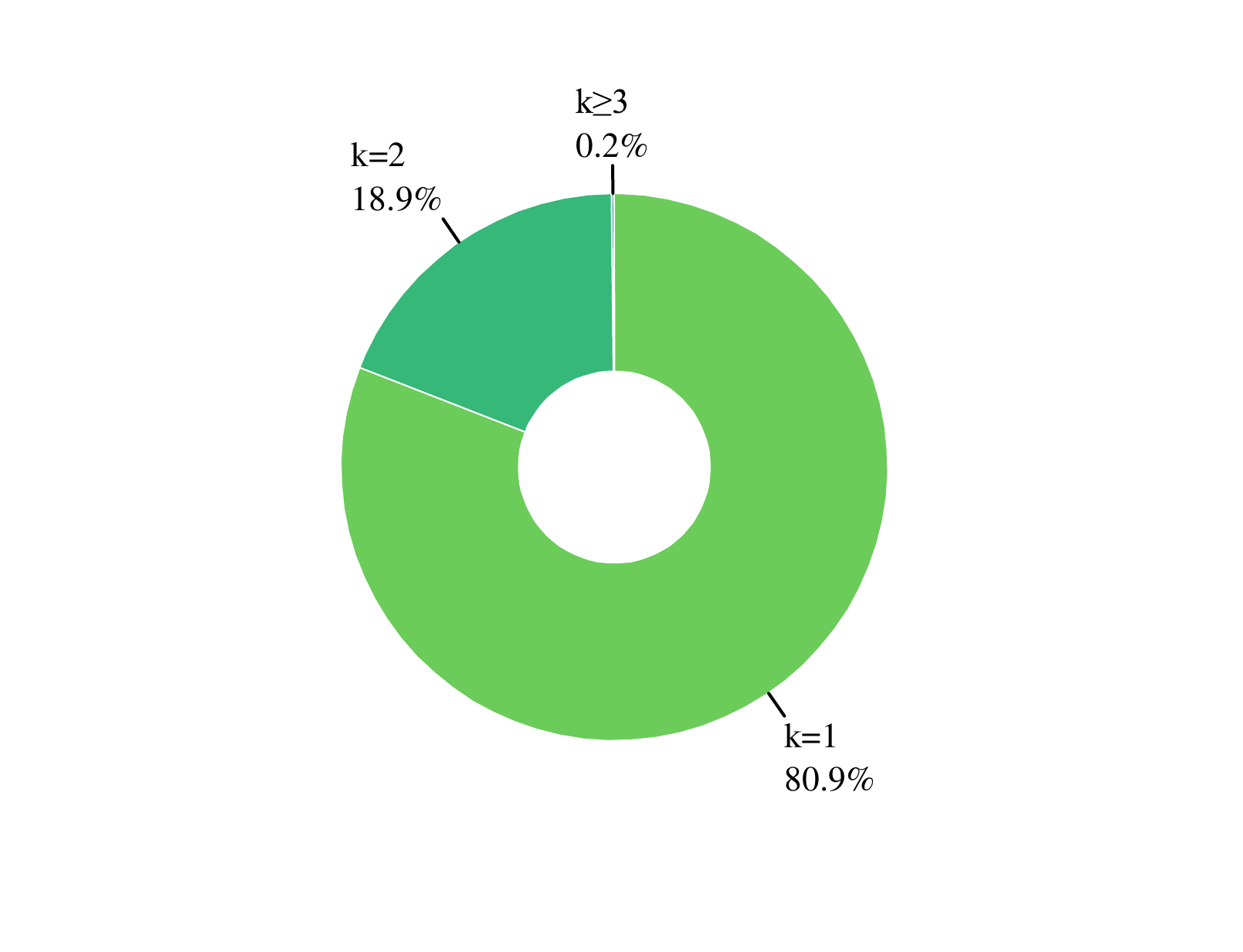}
\put(2,80){\small (a)}
\end{overpic}
\hspace{0.02\linewidth}
\begin{overpic}[width=0.42\linewidth,trim=1cm 0.5cm 2cm 2cm,clip]{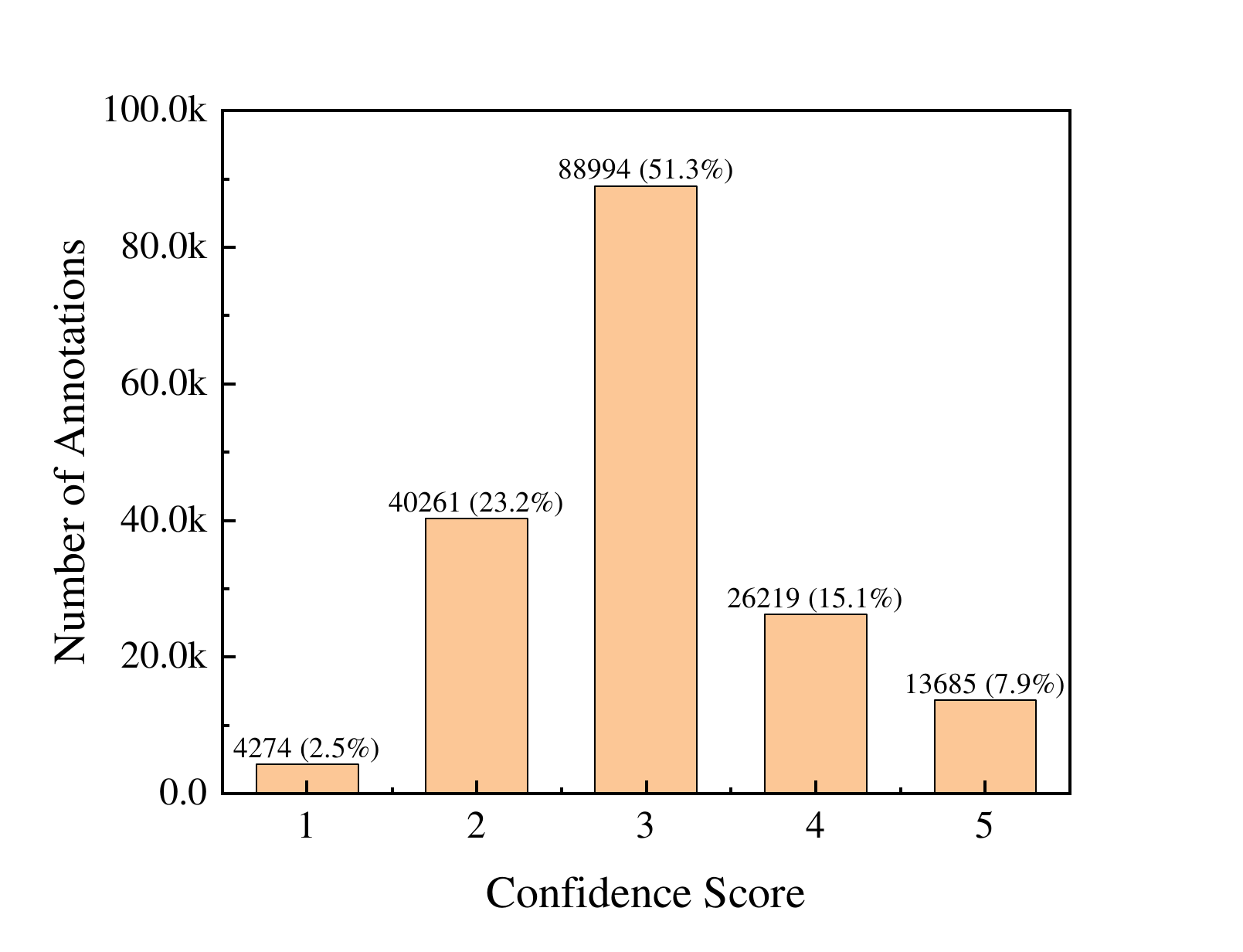}
\put(2,80){\small (b)}
\end{overpic}

\vskip -1.0\baselineskip plus -1fil
\caption{\textbf{Additional statistics of the \textit{EgoScreen-Emotion} dataset.}
(a) Distribution of the number of emotion selections per frame ($k$).
(b) Distribution of annotator confidence scores. }
\label{fig:dataset_statistics}
\vspace{-3mm}
\vskip -1.0\baselineskip plus -1fil

\end{figure}

To further analyze the annotation characteristics of the 
\textit{EgoScreen-Emotion} (ESE) dataset, we present additional statistical 
results in Fig.~\ref{fig:dataset_statistics}.

\vspace{-7mm}

Fig.~\ref{fig:dataset_statistics}(a) shows the distribution of the number of 
emotion selections per frame ($k$). Most frames contain a single dominant 
emotion (80.9\%), indicating that viewers typically exhibit a clear primary 
emotional response when watching movie scenes. Meanwhile, a smaller portion 
of frames contain multiple emotion labels (18.9\% for two emotions and 0.2\% 
for three or more), reflecting that certain scenes may evoke mixed emotional 
reactions. This observation highlights the importance of supporting 
multi-emotion annotations when modeling viewer-level affective responses.
Fig.~\ref{fig:dataset_statistics}(b) presents the distribution of annotator 
confidence scores. Medium-confidence annotations account for the largest proportion, 
while the frequency gradually decreases toward higher certainty levels. Extremely 
uncertain and extremely certain annotations are both relatively rare. This pattern 
is consistent with the general characteristics of human emotion judgment: in most 
cases, annotators can identify a plausible emotional tendency, but only a smaller 
portion of samples evoke highly certain or very strong emotional responses.

\section{Narrative Compression Details}
Emotional responses during movie watching depend not only on the current frame, but also on accumulated narrative memory. However, directly concatenating all historical segment summaries causes the textual context to grow linearly with time, introduces substantial redundancy, and makes the prompt increasingly difficult for the model to use effectively. 
More importantly, human narrative memory is inherently hierarchical: events from the distant past are usually retained as a few key story points, while recently observed events can still be remembered in much greater detail. 
Motivated by this short-term/long-term memory pattern, we compress earlier narrative history into a fixed-schema long-term background, while preserving the most recent segment as a detailed local context.

\subsection{Compression Objective and Schema}

The compressed long-term background is organized around four core dimensions:
\begin{compactitem}
    \item \textbf{Story Premise}: the high-level situation and story setup accumulated so far;
    \item \textbf{Key Entities}: recurring characters, objects, or forces that remain relevant;
    \item \textbf{Conflict}: stable antagonism, danger, or emotional tension that drives the scene;
    \item \textbf{Global Tone}: the dominant affective tendency of the preceding narrative.
\end{compactitem}
In addition, we maintain an auxiliary field \textbf{Turning Points} to record major state changes or salient events that explain why the current scene should be interpreted in a particular way. 
This schema preserves affect-relevant narrative structure while discarding redundant segment-level details, thereby providing a more compact and stable form of narrative memory for emotion prediction.

\subsection{Rolling Compression Procedure}
\label{sec:rolling_compression}

We divide each video into fixed 20-second segments and generate one raw segment summary for each segment, denoted as \(summary_i\), where \(i=1,2,\dots,n\). 
Instead of concatenating all previous summaries, we recursively compress earlier history into a structured long-term background.

To make the compression schema explicit, we represent the compressed background after the first \(i\) completed segments as
\begin{equation}
B_i=\{P_i,E_i,C_i,T_i\},
\end{equation}
where \(P_i\) denotes the accumulated \emph{premise}, \(E_i\) the set of salient \emph{entities}, \(C_i\) the dominant \emph{conflict relations or threats}, and \(T_i\) the \emph{global narrative tone}.

Let the structured information extracted from the current segment summary \(summary_i\) be
\begin{equation}
G(summary_i)=\{\hat{P}_i,\hat{E}_i,\hat{C}_i,\hat{T}_i\},
\end{equation}
where \(\hat{P}_i,\hat{E}_i,\hat{C}_i,\hat{T}_i\) denote the newly identified premise-level description, entities, conflict cues, and tone from the current segment.

The rolling compression is then defined as
\begin{equation}
B_1 = G(summary_1),
\end{equation}
\begin{equation}
B_i = \Psi(B_{i-1},\, G(summary_i)), \quad i\ge2,
\end{equation}
where \(\Psi(\cdot)\) is a schema-aware compression operator that updates each dimension:
\begin{equation}
P_i = \psi_P(P_{i-1}, \hat{P}_i),
\end{equation}
\begin{equation}
E_i = \psi_E(E_{i-1}, \hat{E}_i),
\end{equation}
\begin{equation}
C_i = \psi_C(C_{i-1}, \hat{C}_i),
\end{equation}
\begin{equation}
T_i = \psi_T(T_{i-1}, \hat{T}_i).
\end{equation}

Here, \(\psi_P\) integrates the high-level story setup, \(\psi_E\) retains recurring characters, objects, or forces, \(\psi_C\) updates stable conflict or threat relations, and \(\psi_T\) summarizes the dominant affective tendency of the preceding narrative.

At time step \(t\), let
\begin{equation}
k=\left\lfloor \frac{t}{20}\right\rfloor
\end{equation}
denote the index of the most recently completed 20-second segment. The narrative context is then constructed as
\begin{equation}
\mathbf{S}_t=\Phi(B_{k-1},\,summary_k),
\end{equation}
where \(B_{k-1}\) is the compressed long-term background summarizing earlier history, and \(summary_k\) is the most recent detailed segment summary. When \(k=1\), no long-term background is available yet, and only the current segment summary is used.

This design follows a memory-inspired principle: older events are retained in a low-resolution structured form, whereas the most recent segment remains uncompressed and detailed. As a result, the prompt length remains controlled even for long videos, while the model still receives both global narrative context and local scene-level evidence.

\subsection{Before-vs.-After Example}

To make the compression behavior more concrete, Table~\ref{tab:compression_example} presents an illustrative example using \textit{Avatar} at \(t=160\)s. 
Before compression, all preceding segment summaries are injected as cumulative plain narrative text. 
After compression, earlier history is reorganized into a fixed-schema background, while the current 20-second segment is retained as a detailed narrative description.

\begin{table}[h]
\caption{
Illustrative example of narrative context injection at \(t=160\)s for \textit{Avatar}. 
Compared with cumulative segment summaries, the compressed representation preserves global narrative structure while reducing prompt length by approximately 72\%.
}
\label{tab:compression_example}
\centering
\scriptsize
\setlength{\tabcolsep}{4pt}
\renewcommand{\arraystretch}{1.08}
\begin{tabular}{p{0.46\linewidth} p{0.46\linewidth}}
\toprule
\rowcolor{gray!20}\textbf{Before: cumulative plain narrative} & \textbf{After: structured compressed background + current segment} \\
\midrule

All previous segment summaries from \(0\)--\(160\)s are concatenated as cumulative narrative text. 
For example: ``a wheelchair user appears in a crowded bar \dots soldiers prepare for military action \dots blue-skinned aliens stand on a battlefield while soldiers panic in a control room \dots''. 
This representation preserves local narrative details, but introduces substantial redundancy as the history grows.

&
Earlier history is compressed into a fixed schema:

\textit{story premise}: humans and Na'vi gradually move toward military conflict.

\textit{conflict}: escalating military tension and impending attack.

\textit{key entities}: soldiers, scientists, blue-skinned aliens, aircraft, robots.

\textit{global tone}: tense.

The most recent \(160\)--\(180\)s segment is retained as a detailed narrative description.

\\
\midrule

Approx. \textbf{401 words} (cumulative summaries)
&
Approx. \textbf{112 words} (compressed background + current segment)

\\
\midrule
\multicolumn{2}{c}{\textbf{Word reduction: \(\sim72\%\)}} \\
\bottomrule
\end{tabular}
\vskip -3ex
\end{table}

\subsection{Ablation on Narrative Summary Compression}

To evaluate the effect of narrative summary compression itself, we conduct a text-only ablation that compares two contextual representations: the original narrative summary (\textit{Summary}) and the compressed narrative summary (\textit{Compressed Summary}). In this setting, no additional visual frames or audio inputs are provided; the model receives only the narrative text and is trained to predict the emotion label at the corresponding time step. The first setting directly uses the original narrative summary as input, while the second replaces it with the compressed narrative summary, which reorganizes earlier history into a fixed-length structured representation. All other training configurations remain unchanged, so that the comparison directly isolates the impact of the narrative compression strategy.

\vspace{-3mm}
\begin{table}[h]
\caption{
Effect of narrative summary compression on the 
\textit{EgoScreen-Emotion} test set.
}
\vskip -2ex
\label{tab:summary_compression}
\centering
\small
\setlength{\tabcolsep}{8pt}
\resizebox{.6\textwidth}{!}{
\begin{tabular}{@{}lccc@{}}
\toprule
\rowcolor{gray!20}\textbf{Setting} & \textbf{Acc} & \textbf{Macro-F1} & \textbf{Weighted-F1} \\
\midrule
Summary & 55.50 & 12.12 & 46.61 \\
\textbf{Compressed Summary} & \textbf{59.72} & \textbf{15.82} & \textbf{54.76} \\
\bottomrule
\end{tabular}}
\vskip -3ex
\end{table}

As shown in Table~\ref{tab:summary_compression}, the compressed narrative summary consistently improves performance across all metrics. Accuracy increases from 55.50 to 59.72, Macro-F1 from 12.12 to 15.82, and Weighted-F1 from 46.61 to 54.76. These results suggest that the compressed representation provides a more stable and effective high-level narrative context for emotion prediction. We attribute this improvement to the fact that the original narrative summaries often contain redundant segment-level details that are not always useful for emotion inference. In contrast, the compressed summaries preserve long-term narrative structure in a more compact form, allowing the model to focus more effectively on globally relevant contextual cues.

\section{Annotation Details}
\subsection{Confidence Voting Details}
\begin{figure}[t]
\centering
\includegraphics[width=\linewidth]{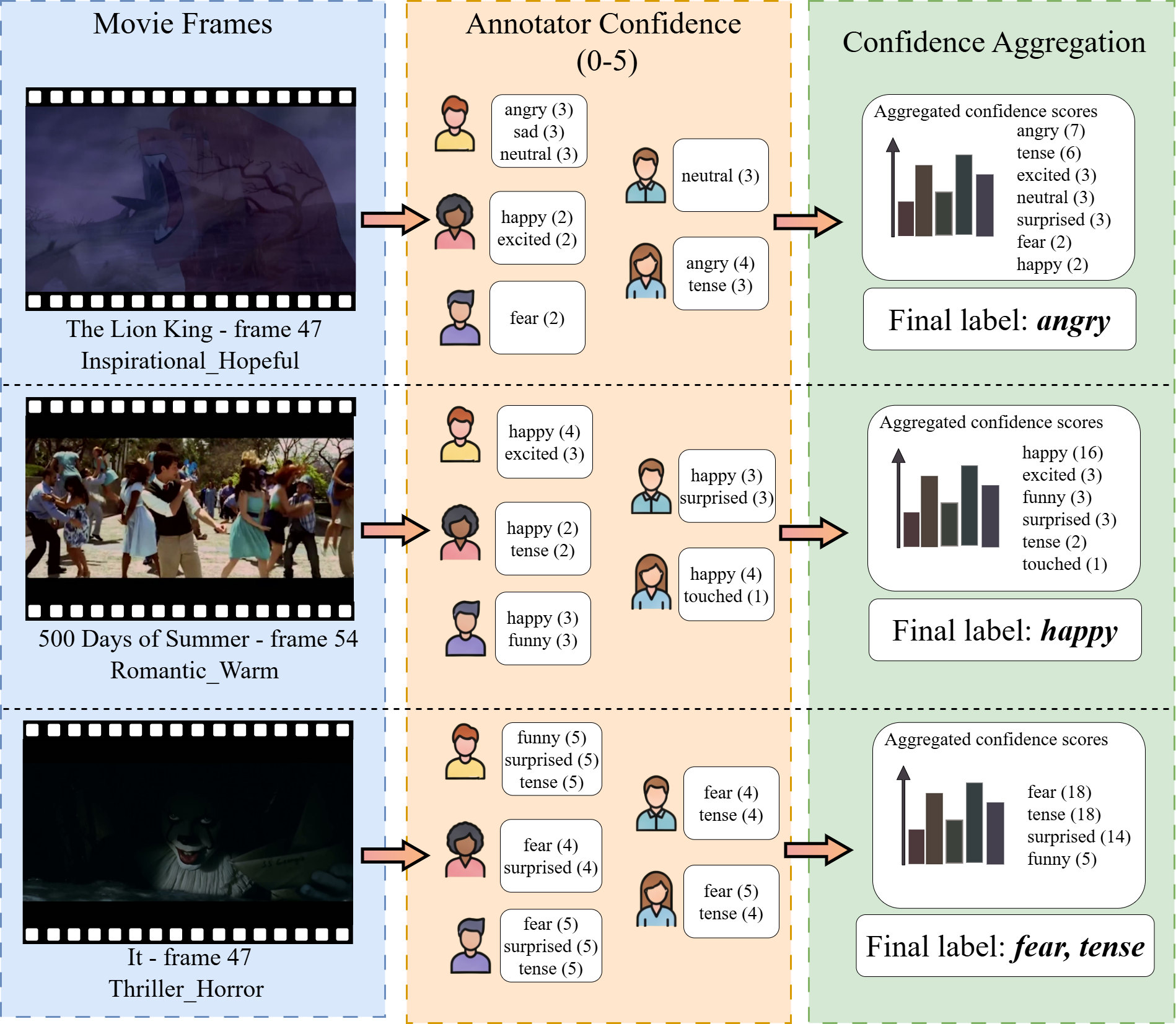}
\caption{
Illustration of the confidence-based emotion aggregation process.
The examples demonstrate three representative scenarios:
dominant emotion, close scores, and tie cases where multiple
emotions may be retained.
}
\label{fig:confidence_voting}
\end{figure}

Fig.~\ref{fig:confidence_voting} illustrates the aggregation process
with three representative scenarios. When one emotion obtains a clearly
higher aggregated confidence score, it is selected as the final label
(\emph{dominant emotion}). When several emotions obtain similar scores,
the emotion with the highest score is still chosen (\emph{close scores}).
If multiple emotions obtain identical highest scores, multiple labels
may be retained (\emph{tie case}).

\vspace{-4mm}

\subsection{Annotation Rationales}

\begin{figure}[t]
\centering
\includegraphics[width=\linewidth]{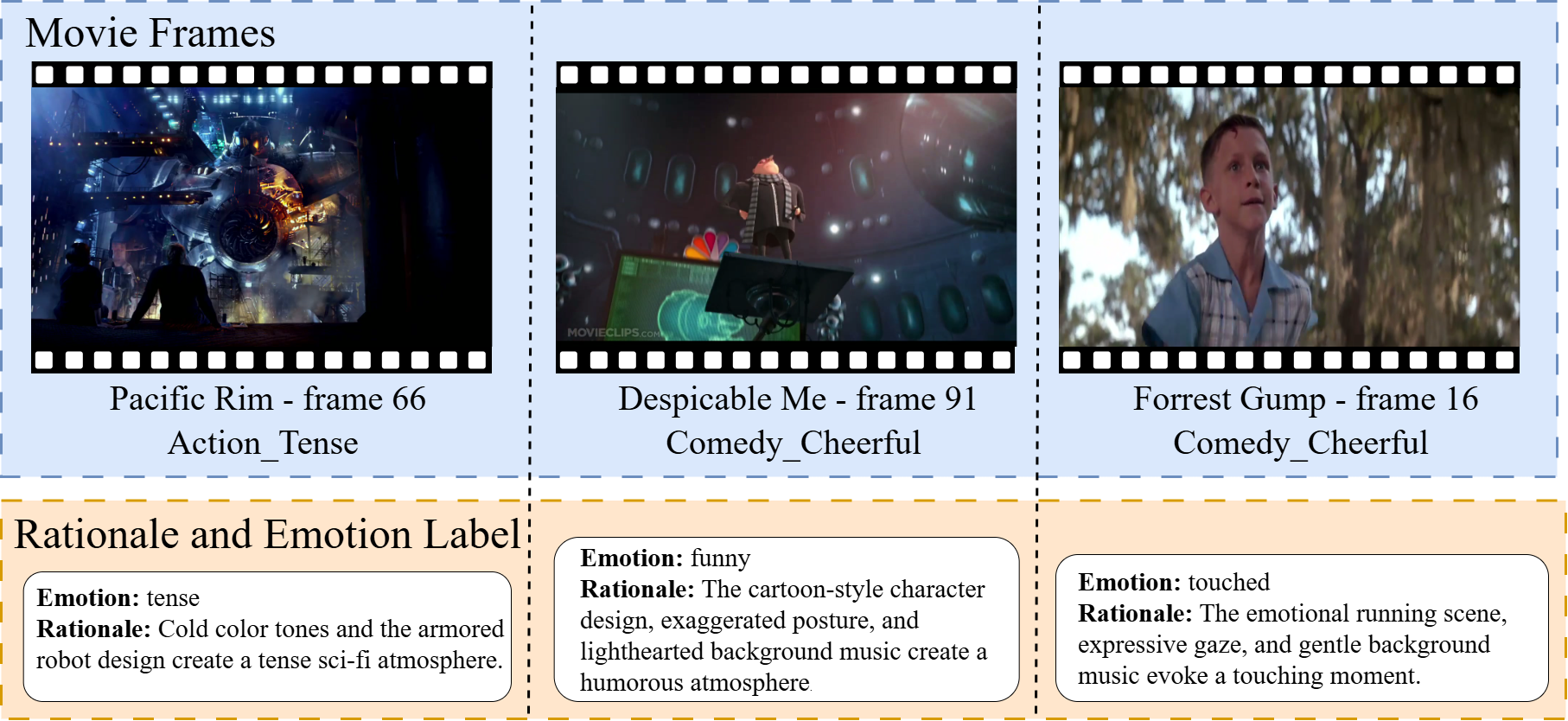}
\caption{
Examples of annotation rationales in the \textit{EgoScreen-Emotion}
dataset. 
}
\label{fig:annotation_rationales}
\end{figure}

Fig.~\ref{fig:annotation_rationales} presents representative examples
of annotation rationales in the \textit{EgoScreen-Emotion} dataset.
During annotation, annotators are required to provide a short textual
explanation describing the reasoning behind their emotional judgment.
These rationales typically summarize the cues or contextual factors
that lead to the perceived emotional response.

\vspace{-3mm}

\section{Prompt and Training Details}
This section provides additional details about the training prompts and implementation settings used in our experiments. 
We first present representative prompt templates for different model configurations, including the single-frame baseline and the multimodal setting with temporal frames, audio, and compressed narrative context. 
We then describe the key training configurations used for model fine-tuning.

\FloatBarrier
\subsection{Prompt}
\begin{mybox}{Single-frame Prompt Example (1F Baseline)}

\footnotesize
\textbf{Training prompt used for the single-frame baseline.}

\begin{verbatim}
{
  "messages": [
    {"role": "user", "content": "<image>

The image above is a frame from a movie clip.
Determine the primary emotion you experience as a viewer
based on the visual content of this frame.

Emotion categories: angry, excited, fear, funny, happy, neutral, sad, 
surprised, tense, touched

Select the single emotion that best describes your response.
Only output the emotion label.

Output format:
{\"emotion\":\"<label>\"}"
    },
    {"role": "assistant", "content": "{\"emotion\":\"happy\"}"}
  ],
  "images": ["<frame_t>"]
}
\end{verbatim}

\end{mybox}

\begin{mybox}{Multimodal Prompt (3F + Audio + Compressed Narrative)}

\scriptsize
\textbf{Training prompt used for the multimodal model.}

\begin{verbatim}
{
  "messages": [
    {"role": "user", "content": "Frame t-10s: <image>
Frame t-5s: <image>
Frame t (current): <image>

Audio segment: <audio>

The frames above are key visual moments from 
the same movie clip arranged in chronological order.

[Narrative Context]
[Compressed Long-Term Background]

story: A humanoid robot in black armor rapidly moves through a parking lot and 
attacks a man using a mechanical arm. The man attempts to escape but is eventually struck.

entities: police officers, robotic threat, man, woman, child

conflict: pursuit, armed confrontation, close combat, vehicle collision, physical attack

tone:tense

[Current Segment: 80–100s]

A woman and a man stand near a body of water and appear to be talking. The scene then 
shifts to a helicopter pilot wearing a reflective vest inside a cockpit filled with
instruments. Later, a person in a dimly lit room holds a black gun while remaining alert.
The scene moves to a prison corridor where a uniformed officer walks while escorting a
woman. Finally, another officer rushes into the prison as if attempting to pursue or 
rescue someone.

Determine the primary emotion experienced by the viewer.

Emotion categories: angry, excited, fear, funny, happy,
neutral, sad, surprised, tense, touched

Output format:
{\"emotion\":\"<label>\"}"
    },
    {"role": "assistant", "content": "{\"emotion\":\"neutral\"}"}
  ],
  "images": ["<frame_t-10>", "<frame_t-5>", "<frame_t>"],
  "audios": ["<audio_segment>"]
}
\end{verbatim}

\end{mybox}

\subsection{Training Details}

We fine-tune Qwen2.5-Omni-7B~\cite{xu2025qwen25omni} and Qwen3-VL-8B~\cite{bai2025qwen3vl} with parameter-efficient LoRA adaptation~\cite{hu2022lora}. 
For all experiments, training is performed for 4 epochs with a learning rate of 1e-4. 
The per-device batch size is set to 1, and gradient accumulation steps are set to 2, resulting in an effective batch size of 2. 
We adopt bfloat16 mixed-precision training and enable gradient checkpointing to reduce GPU memory consumption. 
For Qwen2.5-Omni-7B, we use the SDPA attention~\cite{vaswani2017attention} implementation, while for Qwen3-VL-8B, we use FlashAttention~\cite{dao2022flashattention}. 
Model checkpoints are evaluated and saved at the end of each epoch. 
All experiments are conducted on NVIDIA RTX 5090 GPUs, 
and validation is performed on the corresponding test split of each setting. 
The dataset is split at the movie level to prevent narrative leakage, 
ensuring that clips from the same movie never appear in both the training 
and test sets.

\section{Detailed Evaluation Results}

\subsection{Per-class Performance}

As shown in Table~\ref{tab:per_class_results}, there are clear performance differences across emotion categories. In particular, \textit{neutral} and \textit{tense} achieve substantially higher performance, while \textit{angry} and \textit{touched} obtain F1 scores of 0. This is mainly caused by the long-tailed distribution of the dataset: \textit{neutral} and \textit{tense} account for approximately 53.4\% and 29.3\% of the samples, whereas rare emotions such as \textit{angry} and \textit{touched} constitute only about 1\% and 0.5\%, respectively. As a result, the number of test samples for these rare categories is very limited, making the F1 score sensitive to small prediction changes.

Importantly, this distribution is consistent with the typical emotional response patterns of viewers during movie watching. \textit{Neutral} is a common viewing state, as viewers do not continuously experience strong emotional fluctuations throughout a film, leading to its relatively high proportion in the dataset. Meanwhile, \textit{tense} represents a broad and composite emotional state that can arise in a wide range of narrative situations, such as conflict, suspense, or potential threats, and therefore also appears frequently. In contrast, stronger emotions such as \textit{angry} or \textit{touched} usually emerge only at specific narrative moments and thus occur less frequently overall. Therefore, the long-tailed emotion distribution observed in the ESE dataset reflects a natural pattern of viewer emotional responses rather than an artificial imbalance introduced during dataset construction.

\begin{table}[t]
\centering
\caption{Per-class performance (\%) of the multimodal model 
(3F + audio + compressed narrative) on the ESE test set. 
The last column shows the proportion of each emotion category in the test set.}
\label{tab:per_class_results}
\vspace{-2mm}
\small
\setlength{\tabcolsep}{7pt}
\begin{tabular}{lcccc}
\toprule
\rowcolor{gray!20}
Emotion & Precision (\%) & Recall (\%) & F1 (\%) & Proportion (\%) \\
\midrule
angry      & 0.00  & 0.00  & 0.00  & 1.01 \\
excited    & 24.64 & 16.04 & 19.43 & 1.98 \\
fear       & 40.48 & 10.12 & 16.19 & 3.14 \\
funny      & 24.59 & 17.05 & 20.13 & 1.64 \\
happy      & 33.85 & 44.67 & 38.52 & 4.56 \\
neutral    & 70.33 & 83.29 & 76.26 & 53.41 \\
sad        & 38.20 & 19.10 & 25.47 & 3.32 \\
surprised  & 12.50 & 6.78  & 8.79  & 1.10 \\
tense      & 59.40 & 50.67 & 54.69 & 29.34 \\
touched    & 0.00  & 0.00  & 0.00  & 0.50 \\
\midrule
\multicolumn{5}{c}{Acc 63.01 \quad Macro-F1 25.95 \quad Weighted-F1 60.70} \\
\bottomrule
\end{tabular}
\end{table}

\subsection{Confusion Matrix Analysis}
\begin{figure}[t]
\centering
\includegraphics[width=0.5\linewidth]{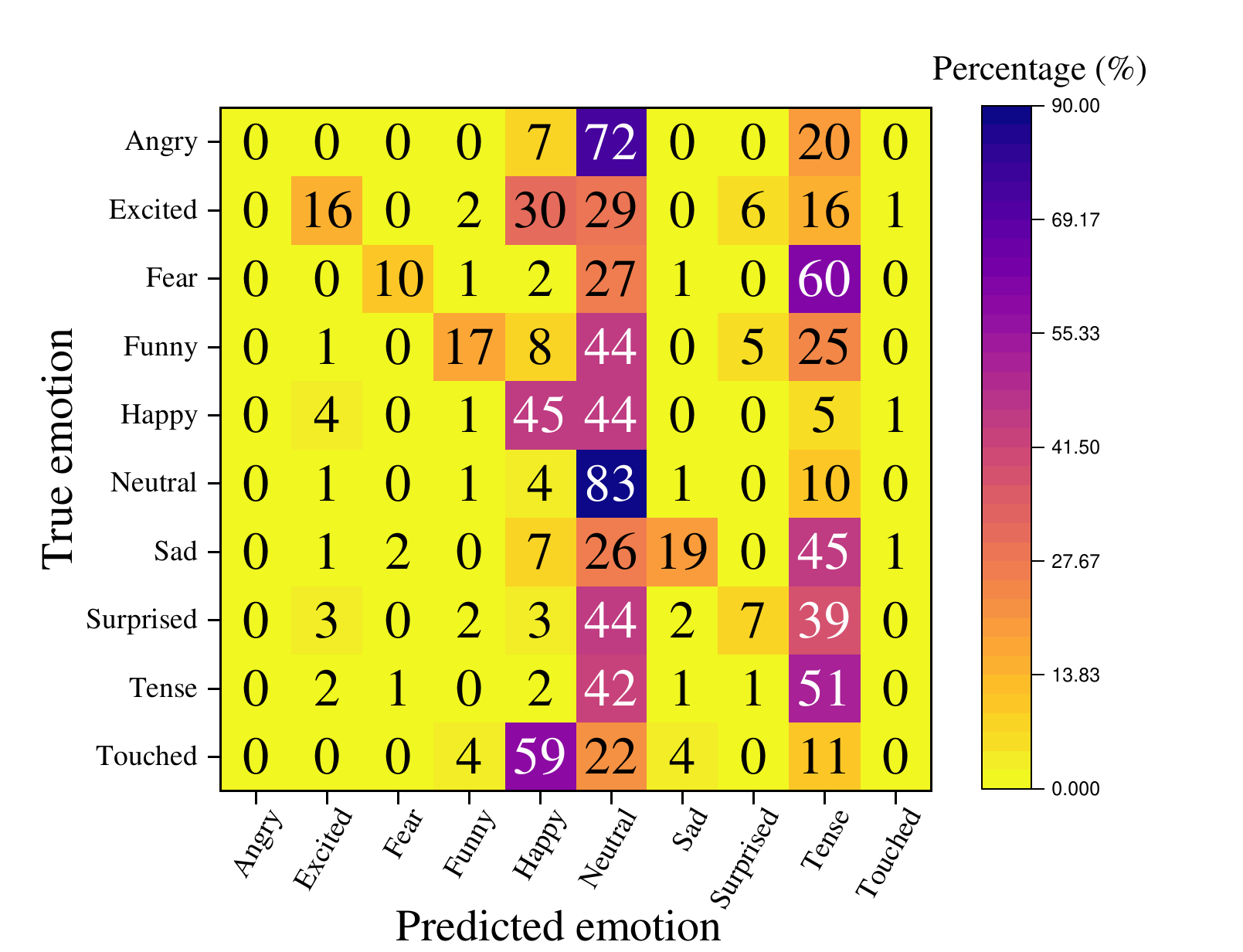}
\caption{
Confusion matrix of the multimodal model on the ESE test set. 
Rows represent true emotion labels and columns represent predicted labels.
}
\label{fig:confusion_matrix}
\end{figure}

Fig.~\ref{fig:confusion_matrix} reveals several meaningful confusion patterns. 
Many emotions are frequently predicted as \textit{neutral}, suggesting that when emotional cues are weak or ambiguous, the model tends to default to a neutral interpretation, which is consistent with viewer-level emotional perception during movie watching.

Fig.~\ref{fig:confusion_matrix} reveals several meaningful confusion patterns. 
Many emotions are frequently predicted as \textit{neutral}, suggesting that when emotional cues are weak or ambiguous, the model tends to default to a neutral interpretation, which is consistent with viewer-level emotional perception during movie watching.

A strong interaction is observed between \textit{fear} and \textit{tense}, where many \textit{fear} samples are predicted as \textit{tense}. This reflects the close semantic relationship between fear and tension in cinematic scenes involving danger or suspense.

Although the F1 scores of \textit{angry} and \textit{touched} are 0 due to their extremely low frequency, the confusion matrix shows that the model still captures their semantic characteristics. For example, \textit{angry} samples are mainly predicted as \textit{neutral} or \textit{tense}, while \textit{touched} samples are often predicted as \textit{happy}. These predictions remain semantically reasonable and fall within the affective neighborhood of the target emotions.

%
%

\end{document}